\documentclass{article} 
\usepackage{iclr2027_conference,times}

\usepackage{amsmath,amsfonts,bm}

\def\eqref#1{equation~\ref{#1}}

\def\1{\bm{1}}

\DeclareMathAlphabet{\mathsfit}{\encodingdefault}{\sfdefault}{m}{sl}
\SetMathAlphabet{\mathsfit}{bold}{\encodingdefault}{\sfdefault}{bx}{n}

\usepackage{hyperref}
\usepackage{url}
\usepackage{fontawesome5}

\usepackage{wrapfig}

\usepackage[utf8]{inputenc}
\usepackage{verbatim}
\usepackage{calc}
\usepackage{mathrsfs}
\usepackage{mathtools}
\usepackage{url}
\usepackage{algorithm}
\usepackage{algpseudocode}
\usepackage{tikz}
\usetikzlibrary{arrows.meta,calc,positioning,shapes.geometric}
\usepackage{amsmath}
\usepackage{amssymb}
\usepackage{graphicx}
\usepackage{hyperref}
\usepackage{xcolor}
\usepackage{comment}
\usepackage{soul}
\usepackage{booktabs}
\usepackage{xspace}

\definecolor{forestgreen}{rgb}{0.13, 0.55, 0.13}

\usepackage{tabularx}
\usepackage{multirow}
\usepackage{multicol}
\usepackage{subcaption}
\usepackage[most]{tcolorbox}

\definecolor{curatorpromptborder}{HTML}{14233F}
\definecolor{curatorpromptheader}{HTML}{E8EEF5}
\definecolor{curatorpromptbody}{HTML}{F7F9FC}
\newtcolorbox{curatorprompt}[1]{
    enhanced,
    breakable,
    colback=curatorpromptbody,
    colframe=curatorpromptborder,
    colbacktitle=curatorpromptheader,
    coltitle=curatorpromptborder,
    fonttitle=\bfseries,
    fontupper=\footnotesize,
    title={#1},
    boxrule=0.4pt,
    titlerule=0.4pt,
    arc=1.5pt,
    outer arc=1.5pt,
    left=7pt,
    right=7pt,
    top=6pt,
    bottom=6pt,
    before skip=7pt,
    after skip=7pt
}

\usepackage{colortbl}
\usepackage{tabulary}
\usepackage{etoolbox}

\definecolor{colorA}{RGB}{189,201,225}
\definecolor{colorB}{RGB}{103,169,207}
\definecolor{colorC}{RGB}{ 28,144,153}
\definecolor{colorD}{RGB}{  1,108, 89}

\newcolumntype{R}{>{\columncolor{gray!40}}r}
\newcolumntype{L}{>{\columncolor{gray!40}}l}
\newcolumntype{C}{>{\columncolor{gray!40}}c}

\newcommand{\graycell}[1]{\cellcolor{teal!15}#1}

\newcommand{\OURS}{GEPA-Curator\xspace}

\usepackage{microtype}
\usepackage{graphicx}
\usepackage{booktabs} 

\usepackage{enumitem}

\newcommand\appref{Appendix~\ref}

\newcommand\fref{Figure~\ref}
\newcommand\tref{Table~\ref}
\newcommand\sref{Section~\ref}

\usepackage{amssymb}
\usepackage{pifont}
\definecolor{darkerred}{RGB}{190, 0, 0} 

\title{GEPA-Curator: \\ Curated Learning for Self Improving Agents}

\author{%
\begin{minipage}[t]{\textwidth}
\centering\normalfont\bfseries
Zack Yu\textsuperscript{1*},\enskip
Nicholas Lee\textsuperscript{1*},\enskip
Minjae Lee\textsuperscript{4},\enskip
Lakshya A Agrawal\textsuperscript{1},\enskip
Luca Manolache\textsuperscript{1}
\\[0.5mm]
Wonjun Kang\textsuperscript{4},\enskip
Matei Zaharia\textsuperscript{1},\enskip
Michael W. Mahoney\textsuperscript{1,2,3},\enskip
Kurt Keutzer\textsuperscript{1}
\\[0.5mm]
Sehoon Kim\textsuperscript{5$\dagger$},\enskip
Amir Gholami\textsuperscript{1,2$\dagger$}
\\[2mm]
{\small\mdseries
\textsuperscript{1}UC Berkeley
\enspace
\textsuperscript{2}ICSI
\enspace
\textsuperscript{3}LBNL
\enspace
\textsuperscript{4}Furiosa AI
\enspace
\textsuperscript{5}KAIST}
\\[1.5mm]
\end{minipage}%
}

\iclrfinalcopy 
\begin{document}

\maketitle
\begingroup
\renewcommand{\thefootnote}{}
\footnotetext{
$^{*}$Equal contribution.
$^{\dagger}$Equal advising.
Correspondence to: Amir Gholami (amirgh@berkeley.edu). \\
} 
\endgroup
\raggedbottom

\begin{abstract}
Large Language Models (LLMs) are increasingly used in multi-turn agentic workloads that require sustained interaction with complex environments.
Existing reflective prompt optimizers
such as GEPA learn from failed rollouts, but agentic failures can also come from prompt weaknesses, harness limitations, stochastic execution, or noisy judges. 
We argue that such failures may not provide useful learning signal, and may even be harmful; and thus it is essential to curate which tasks are provided to the prompt optimizer for better performance. 
We introduce \OURS, a framework that augments the prompt optimizer to select which agent experiences should shape prompt optimization. 
In particular, \OURS consists of two components: \emph{task curation}, which filters out tasks that are likely to provide noisy or unreliable learning signals, and \emph{prior success retrieval}, which selectively retrieves past successful trajectories for a task and provides them as additional context to the prompt optimizer.
We evaluate \OURS
on GoBrowse, ALFWorld, AppWorld, and $\tau$-bench, using both Gemini 3.1 Pro and Claude Opus 5 as optimizer models. 
\OURS matches or exceeds the mean performance of both baselines, with gains of up to 20.7\% over GEPA and 21.2\% over ACE. 
\end{abstract}
\section{Introduction}

As interest in recursive self-improvement (RSI) grows, a central question is how agents can reliably learn from their own experience to improve future behavior~\citep{anthropic2026recursive}. 
Reflective prompt optimization, where an agent acts with tools in an environment and a Reflector uses the resulting trajectory or end state to identify lessons and revise the agent's prompt, offers one practical mechanism for this form of self-improvement~\citep{agrawal2026gepa, zhang2026agentic}.
The revised prompt is then used on subsequent tasks, creating a feedback loop in which experience can progressively improve agent behavior. 
In agentic settings, this approach rests on the assumption that experience from previously attempted tasks contains signal and information that generalize to future tasks.

However, it is often difficult for the Reflector to correctly attribute the cause of a failure to the prompt.
In agentic tasks, execution is mediated by a harness interacting with an external environment, and success is often assessed by LLM-as-a-judge~\citep{zheng2023judging}. 
A failed rollout may therefore reveal a genuine prompt weakness, but it may also arise from external factors such as unavailability of needed tools, harness limitations, or even noise from execution or evaluation environments.
This problem is further exacerbated in real-world settings, where tasks themselves may be ambiguous or infeasible, introducing additional noisy signals to the Reflector.
When revising the prompt, the Reflector may nevertheless treat these potentially noisy observations as valid learning signals for \textit{prompt} optimization~\citep{kirichenko2025abstentionbench}, causing it to misattribute the associated failures to prompt deficiencies, thereby producing unnecessary or harmful updates.

As an example, consider an agent asked to purchase an item from a shopping website. 
If the agent fails because it overlooks a required option before adding the item to the cart, the Reflector can usefully revise the prompt to encourage checking required fields before proceeding.
In contrast, if the same task fails because it runs in a broken environment where the purchase button in the webpage is temporarily malfunctioning, treating this failure as evidence about the prompt may lead the Reflector to introduce unnecessary or even harmful changes.
For example, it may conclude that 
the agent needs to seek alternative ways to complete the purchase, while the button can become functional for future~tasks.

To mitigate such issues, recent work \citep{agrawal2026gepa, yang2026skillopt} uses a held-out validation set in order to guard against these changes.
If the prompt does poorly on the validation set after mutation, the change can be rejected.
GEPA \citep{agrawal2026gepa} goes a step further and evaluates the mutated prompt on the minibatch again to make sure that the model performance does not degrade after such a mutation.
However, such validation-based gating mechanisms alone may not be sufficient to prevent noisy signals, for two reasons.
First, bad mutations may be accepted and good ones rejected, e.g., because the gating decision relies on \textit{aggregated} evaluation scores, which can be noisy due to stochasticity in execution or evaluation.
Second, such validation-based gating acts only \textit{after} a mutation has been generated, and therefore cannot fundamentally improve the quality of the reflection or mutation itself.
This motivates us to maintain and provide additional contextual information in order to improve the capabilities of the Reflector.

Based on these insights, we present GEPA-Curator, an agentic data Curator which runs as a harness on top of existing prompt optimizers.
GEPA-Curator has two roles. 
The first is \textit{task curation}, where instead of randomly selecting tasks for the minibatch, the agent will curate a minibatch which is likely to provide clean signal and discard tasks which seem to be noisy or faulty.
The second is a \textit{prior success retrieval}, where the Curator can selectively retrieve past successful trajectories for a task and provide them as context to the prompt optimizer to improve the quality of the mutation.
Providing the Reflector agent with successful exemplars is essential to give the model the proper context to see why the prompt is failing now, and when it succeeded in the past.

Overall, our contributions are as follows:

\begin{itemize}[leftmargin=*,nosep]
    \item 
    We introduce \OURS, an agentic harness for prompt optimization that improves mutation quality through adaptive task curation (\sref{sec:task_curation}) and selective retrieval (\sref{sec:success_retrieval}) of prior successful trajectories.
    \item 
    We evaluate \OURS on four WebArena websites \citep{zhou2024webarena}, ALFWorld, AppWorld, and $\tau$-bench. \OURS matches or exceeds the mean performance of the evaluated baselines, with gains of up to 20.7\% over GEPA \citep{agrawal2026gepa} and 21.2\% over ACE \citep{zhang2026agentic}. These results suggest that the benefits of curation extend across optimizer models.
\end{itemize}

\section{Related Work}
\label{sec:related_work}

\paragraph{LLM Agents.}
As LLMs become ubiquitous, LLM-powered agents have been equipped with tools and harnesses in order to interact and execute in external environments. 
React \citep{yao2023react} interwove reasoning with tool execution to improve performance.
Later work \citep{patil2024gorilla, erdogan2024tinyagent, schick2023toolformer, shen2023hugginggpt, xu2023rewoo, kim2024an} used tool calling, where the model generates a structured output which is parsed and used to execute functions to affect the environment.
This has led to LLM agents being used to control and interact with different environments; and 
this has led to a variety of different benchmarks \citep{trivedi2024appworld, shridhar2021alfworld, yao2024tau, patil2024gorilla, patil2025bfcl} to evaluate LLMs at tool calling.

One example is using LLMs as webagents in order to explore and act in web environments~\citep{qi2025webrl, erdogan2025planandact, gur2023understanding, kim2023language, yang2025agentoccam}.
This has led to the development of various browser and computer use benchmarks and environments in this area~\citep{zhou2024webarena, gandhi2026go, he2024webvoyager, xue2025an, xie2024osworld, yuan2026osworld2, rawles2023androidinthewild}.
Webarena \citep{zhou2024webarena} was one of the first web environments and benchmarks to cover realistic websites such as GitLab, Reddit, Shopping portals, etc.

\paragraph{Prompt Optimization.}
Reflective optimizers use natural language feedback in place of numerical gradients to revise textual components of an AI system~\citep{yuksekgonul2024textgrad, agrawal2026gepa, yang2026skillopt, zhang2026agentic, zhou2023large, yang2024large, pryzant2023automatic, opsahl2024optimizing, zhao2024expel}. 
In particular, GEPA \citep{agrawal2026gepa} applies reflection to system-level trajectories, using evaluator feedback to diagnose failures and propose prompt mutations within an evolutionary search.
ACE \citep{zhang2026agentic} introduced a framework for agentic context adaptation to create structured playbooks to organize and accumulate strategies at runtime.
SkillOpt \citep{yang2026skillopt} similarly treats an agent's skill document as trainable external state; a separate optimizer analyzes scored rollouts and proposes bounded add, delete, or replace edits.
Meta-Harness \citep{lee2026metaharness} extended this idea further by having an outer optimization loop on the harness code which runs the agent instead of just the prompt.
These methods demonstrate that execution traces can provide substantially richer optimization signals than scalar rewards alone. 
\OURS addresses a complementary question: \textit{which agentic trajectories provide a valid signal for reflection, and what historical evidence should accompany them?}

\paragraph{Curriculum Learning.}
Curriculum Learning \citep{10.1145/1553374.1553380, fan2018learning} studies how the selection and ordering of training data can improve optimization and model training.
Recent work \citep{zhou2023lima, chen2024alpagasus, wang2024data, lee2024llm2llm, diao2024active} has looked into how this can affect LLM training and alignment. 
LIMA \citep{zhou2023lima} established how having poorer quality data can negatively affect LLM alignment, and showed that a small amount of high-quality data can be as effective as a large scale training set.

\section{\OURS Methodology}
\label{sec:method}

\begin{figure}[!t]
    \centering
    \includegraphics[width=\linewidth]{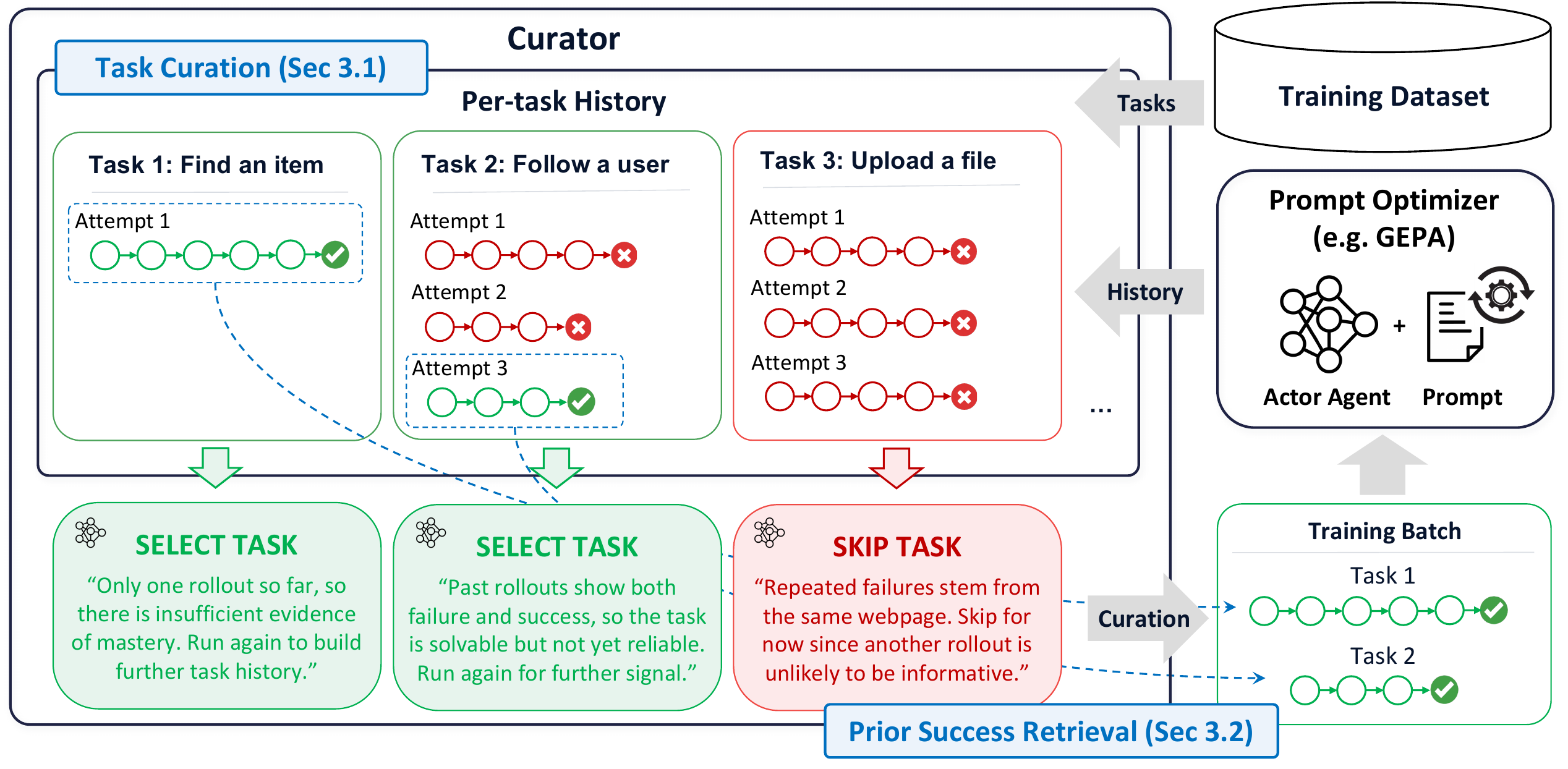}
    \caption{
    \OURS system diagram. 
    The prompt optimizer takes in a minibatch of tasks at each iteration and invokes the actor agent to execute them and performs reflective prompt optimization based on the resulting trajectories. 
    The Curator augments the prompt optimizer by acting as an outer-loop process which traverses through the training dataset and constructs each minibatch to pass into the prompt optimizer. 
    The Curator reads past execution trajectories for each task sampled from the training dataset and decides to run the task if doing so is likely to provide useful optimization signal (\sref{sec:task_curation}). 
    The Curator can also pass prior successful trajectories for failed tasks as context to the prompt optimizer as a trusted reference example (\sref{sec:success_retrieval}). 
    Once the curated training batch of selected tasks and successful context is constructed, it is passed to the prompt optimizer for execution and optimization. 
    The resulting trajectories and evaluation outcomes are then written back to the task history for use in subsequent curation rounds. 
    }
    \label{fig:curriculum-Curator-workflow}
\end{figure}

In this section, we describe how \OURS integrates with existing prompt optimizers.
To do so, we first formulate a prompt optimizer by its input-output interface. 
At each step, a prompt optimizer receives a prompt, a minibatch of tasks, and optionally some input metadata. It executes the tasks using an actor agent conditioned on the prompt and returns an updated prompt, task-level agent trajectories, and evaluation results (See \fref{fig:curriculum-Curator-workflow}).
Ordinary prompt optimizers randomly permute and iterate through the tasks in the training dataset at each step.
We identify a few points in existing prompt optimizers where noisy optimization signals can arise.
When selecting tasks to execute, some tasks may yield noisy or unhelpful trajectories.
Furthermore, when reflecting on execution results, some failures may be difficult to attribute correctly from the failed trajectory alone. 
In \sref{sec:motivating-examples}, we present motivating experiments showing how such noisy signals can propagate through the optimization process and ultimately degrade training.

\OURS runs as a harness on top of the optimizer, 
maintaining a per-task history, including agent trajectories and evaluation outcomes from previous optimization steps.
The Curator uses this history to address these sources of noise through two corresponding roles:
(i) \textit{task curation} (\sref{sec:task_curation}), which selects tasks likely to provide useful signal for prompt optimization to construct the next minibatch;
and 
(ii) \textit{prior success retrieval} (\sref{sec:success_retrieval}), which retrieves relevant past successful trajectories for tasks in the current minibatch to provide the Reflector with concrete contextual references for failure diagnosis.
Note that these trajectories are provided to the prompt optimizer only as additional input metadata, leaving the optimizer itself unchanged.
The full workflow is depicted in \fref{fig:curriculum-Curator-workflow}.
We have found that both of these methods are critical for performance (\sref{sec:ablations}).

\subsection{Task Curation}
\label{sec:task_curation}

The Curator examines the training tasks and selects which tasks to run at each optimization step.
Underexplored tasks are initially prioritized to build sufficient per-task history. 
As this history accumulates, the Curator may skip a task when there is strong evidence that further rollouts are unlikely to yield useful reflection signal for prompt optimization.
Such cases include repeated stable success, repeated failures caused by factors outside the prompt (e.g., harness or environment limitations), and inconsistent evaluations of similar outcomes, which can be judged based on the accumulated per-task history.

Importantly, skipping is a temporary decision rather than a permanent decision, and thus skipped tasks are reconsidered by the Curator at each training epoch.
This is necessary because tasks that were previously solved may regress as the prompt is updated for other tasks, e.g., through prompt compression or the addition of potentially conflicting rules.
Conversely, tasks that were initially too difficult, and therefore skipped after repeated failures, may become solvable as the prompt improves and can be reintroduced into training.

\subsection{Prior Success Retrieval}
\label{sec:success_retrieval}

Normally, prompt optimizers reflect on failure examples in the current minibatch and mutate the prompt based only on the failed trajectories and their evaluations. 
Therefore, pairing a current failed execution with prior successful executions of the same task allows the Reflector to perform \textit{contrastive learning} of the failed behavior against a known and trusted reference.
This provides the Reflector with a concrete basis for correcting the behavior, rather than forcing it to diagnose the failure by guessing.

The Curator is instructed to reason over the task history to select one successful trajectory that provides the most reliable reference for the current task.
In particular, it favors successes supported by consistent evaluation evidence and avoids using outcomes whose apparent success is ambiguous or inconsistent with other observations in the history. 
When no appropriate success is available, the Curator abstains and does not provide this metadata to the Reflector.
This approach prevents noisy success trajectories from providing misleading positive examples, in the same way that removing noisy failure trajectories prevents misleading optimization signals. 
The prompt for task curation and prior success retrieval is provided at Appendix~\ref{appendix:Curator-input}.
Note that because task curation and prior-success retrieval are performed jointly in a single Curator call, the latter requires no additional LLM invocation. Appendix~\ref{app:task-selection-details} analyzes the Curator's task-selection and success-selection decisions.

In summary, these two components of \OURS (task curation and prior success retrieval) work in unison to ensure that the prompt optimizer gets cleaner and curated data to train on.
By curating the tasks in each minibatch, we maximize the amount of signal provided at each step of the optimization loop through removing noisy or redundant tasks.
By providing reliable past successful trajectories to the Reflector, we enable the Reflector to create better mutations with grounded understanding instead of guessing the fix.

\section{Empirical Evaluation}

\subsection{Experimental Setup}
\label{sec:setup}

We compare \OURS against GEPA \citep{agrawal2026gepa} and ACE \citep{zhang2026agentic}. 
\OURS builds on a simplified, greedy variant of GEPA that removes Pareto-based candidate selection. 
Rather than evolving a candidate pool and sampling from the Pareto frontier on validation set, \OURS maintains a single lineage in which each accepted mutation becomes the next checkpoint. 
We evaluate the prompt every 16 iterations unless specified on a held-out validation set and select the final model based on the best validation score at the end.
Because our ablation showed little difference between GEPA with and without Pareto selection, we adopt this greedy variant for simplicity. 

The prompt-optimization experiments use gpt-oss-120b
\citep{agarwal2025gpt} as the actor agent. 
We evaluate two optimizer models, Gemini 3.1 Pro Preview \citep{gemini31pro2026} and Claude Opus 5 \citep{claudeopus52026}, using the selected model for all optimizer-side agents, including the Reflector and Curator, within each run. The minibatch size is 16.
The actor's maximum context length is set to gpt-oss-120B's native maximum context length (131072) so we constrain the optimized prompt to fit within this limit.
All seed prompts used for optimization are manually written.
The evaluation numbers are averaged over 5 repeated runs. Serving details, seed prompts, and judge prompts are in \appref{app:implementation_details}.

For evaluation of optimization on noisy data, we experimented with tasks from GoBrowse \citep{gandhi2026go}: GitLab (720 tasks), Shopping (585 tasks), Shopping Admin (750 tasks), and Reddit (585 tasks) in the WebArena environment \citep{zhou2024webarena}. 
Each website is independently split $3{:}1{:}1$ into train, validation, and held-out test sets. All methods use the same seed prompt, evaluator, action limits, and environment configuration. 
Each website/method uses 5,000 rollouts, which has been chosen to ensure that all the methods can reach their saturated success rate.

We additionally evaluated on more agentic datasets, namely ALFWorld \citep{shridhar2021alfworld}, AppWorld \citep{trivedi2024appworld}, and $\tau$-bench \citep{yao2024tau}. 
For ALFWorld, we sample 150 training tasks and 50 validation tasks from official train and valid\_seen splits to train with 2000 rollouts, and evaluate on all 134 valid\_unseen tasks. 
For AppWorld, we use the official dataset splits with 1500 rollouts in training, and report score based on Task Goal Completion. For $\tau$-bench, we remove user interaction as done in \citep{sayana2026prompting} by flattening all 115 interactive scenarios from the retail test set into single requests, and partition them into 48 training tasks, 32 validation, and 35 held-out test tasks with 800 rollouts. The rollout budget per benchmark was chosen to ensure that all methods can reach their saturated success rate. $\tau$-bench specifically validates every 4 iterations due to its small training data.

\subsection{Main Results}

\paragraph{WebArena Experiments.}
\label{sec:results}

With Gemini 3.1 Pro as the optimizer model, as shown in \tref{tab:webarena-test}, GEPA improves over the seed prompt by 6.3\%, whereas \OURS improves by 9.7\%.
\OURS improves mean held-out test success over GEPA on all four websites, with gains ranging from 1.6-7.4\%. The aggregate difference between the two optimized prompts is 3.4\%. Based on the displayed aggregate scores, \OURS achieves $1.54\times$ GEPA's optimization gain over the seed. This suggests that curation materially increases how much performance can be improved from the same initial prompt.

With Claude Opus 5 as the optimizer model, as shown in \tref{tab:webarena-test}, GEPA improves over the seed prompt by 6.4\%, whereas \OURS improves by 11.4\%. \OURS improves mean held-out test success over GEPA on all four websites, with gains ranging from 0.6\%-7.4\%. The aggregate difference between the two optimized prompts is 5.0\%. Based on the displayed aggregate scores, \OURS achieves $1.78\times$ GEPA's optimization gain over the seed.

Taken together, across both optimizer models and all WebArena domains, \OURS consistently outperforms GEPA, with aggregate improvements of 3.4\% and 5.0\% (and up to 7.4\% for both) for Gemini 3.1 Pro and Claude Opus 5, respectively. This consistency across tasks and optimizer model choices suggests that the benefits of curation are robust rather than specific to a particular run, domain or optimizer. The final optimized prompts and a characterization of how they change during optimization are in \appref{appendix:prompt-mutation}. We further audit the Curator's decisions in \appref{app:evidence-conditioned-audit}.

\begin{figure}
\centering
    \renewcommand{\arraystretch}{1.3}
    \resizebox{\linewidth}{!}{%
    \begin{tabular}{llccccc}
        \toprule
        Optimizer Model & Method & GitLab & Shopping & Shopping Admin & Reddit & Overall \\
        \midrule
        \multirow{3}{*}{Gemini 3.1 Pro} & Seed & $78.2$ {\scriptsize\textcolor{gray}{$\pm$ 2.5}} & $73.5$ {\scriptsize\textcolor{gray}{$\pm$ 11.8}} & $65.2$ {\scriptsize\textcolor{gray}{$\pm$ 6.2}} & $72.6$ {\scriptsize\textcolor{gray}{$\pm$ 2.2}} & $72.4$ {\scriptsize\textcolor{gray}{$\pm$ 2.4}} \\
         & GEPA & $82.1$ {\scriptsize\textcolor{gray}{$\pm$ 2.7}} & $83.8$ {\scriptsize\textcolor{gray}{$\pm$ 2.5}} & $71.5$ {\scriptsize\textcolor{gray}{$\pm$ 2.0}} & $77.6$ {\scriptsize\textcolor{gray}{$\pm$ 2.3}} & $78.7$ {\scriptsize\textcolor{gray}{$\pm$ 0.6}} \\
         & \graycell{\OURS~} & \graycell{$\mathbf{84.6}$ {\scriptsize\textcolor{gray}{$\pm$ 1.5}}} & \graycell{$\mathbf{85.8}$ {\scriptsize\textcolor{gray}{$\pm$ 3.5}}} & \graycell{$\mathbf{73.1}$ {\scriptsize\textcolor{gray}{$\pm$ 1.9}}} & \graycell{$\mathbf{85.0}$ {\scriptsize\textcolor{gray}{$\pm$ 1.8}}} & \graycell{$\mathbf{82.1}$ {\scriptsize\textcolor{gray}{$\pm$ 1.7}}} \\
        \midrule
        \multirow{3}{*}{Claude Opus 5} & Seed & $79.6$ {\scriptsize\textcolor{gray}{$\pm$ 2.3}} & $73.9$ {\scriptsize\textcolor{gray}{$\pm$ 2.5}} & $67.7$ {\scriptsize\textcolor{gray}{$\pm$ 2.9}} & $71.3$ {\scriptsize\textcolor{gray}{$\pm$ 1.0}} & $73.1$ {\scriptsize\textcolor{gray}{$\pm$ 1.8}} \\
         & GEPA & $82.5$ {\scriptsize\textcolor{gray}{$\pm$ 1.8}} & $80.5$ {\scriptsize\textcolor{gray}{$\pm$ 3.2}} & $72.9$ {\scriptsize\textcolor{gray}{$\pm$ 2.4}} & $81.9$ {\scriptsize\textcolor{gray}{$\pm$ 1.5}} & $79.5$ {\scriptsize\textcolor{gray}{$\pm$ 1.2}} \\
         & \graycell{\OURS~} & \graycell{$\mathbf{83.1}$ {\scriptsize\textcolor{gray}{$\pm$ 3.9}}} & \graycell{$\mathbf{86.5}$ {\scriptsize\textcolor{gray}{$\pm$ 3.9}}} & \graycell{$\mathbf{80.3}$ {\scriptsize\textcolor{gray}{$\pm$ 2.9}}} & \graycell{$\mathbf{88.2}$ {\scriptsize\textcolor{gray}{$\pm$ 1.6}}} & \graycell{$\mathbf{84.5}$ {\scriptsize\textcolor{gray}{$\pm$ 1.0}}} \\
        \bottomrule
    \end{tabular}%
    }
    \captionof{table}{Comparison of the success rate (\% with std) on 4 different domains (GitLab, Shopping, Shopping Admin, and Reddit) of the GoBrowse benchmark on the WebArena environment. 
    We evaluate using seed prompt, as well as the prompt optimized by GEPA, and \OURS over held-out test set.
    \OURS exceeds the success rate of all baseline methods across the evaluated settings.
    }

    \label{tab:webarena-test}
\end{figure}

\paragraph{Additional Agentic Benchmarks.}

The results of the additional agentic benchmarks are shown in \tref{tab:agentic-benchmarks-test}. With Gemini 3.1 Pro as the optimizer model, \OURS improves over both baselines across all four test sets, with average gains of 14.3\% over GEPA and 5.0\% over ACE. These results suggest that task curation and prior success retrieval are general methods that can benefit optimization beyond noisy datasets like GoBrowse. 

With Claude Opus 5 as the optimizer model, \OURS improves over GEPA by 2.6\% on ALFWorld, 15.9\% on AppWorld Test-N, and 13.9\% on AppWorld Test-C, while matching its mean success on $\tau$-bench.
Averaged across the four test sets, \OURS gains 8.1\% over GEPA and 12.6\% over ACE. Together, the Gemini and Opus results suggest that the benefits of curation extend across optimizer models, with magnitude depending on the benchmark and optimizer model. Additional evaluation on QA-type benchmarks from \citep{agrawal2026gepa} can be found in \appref{appendix:gepa-qa-benchmarks}. We also compare \OURS with ExpeL \citep{zhao2024expel} in \appref{app:expel-comparison} and post-rollout filtering in \appref{appendix:rollout-filtering}. Self improvement with strictly the same model is evaluated in \appref{app:same-model-self-improvement}.

\begin{figure}[!t]
\centering
   \renewcommand{\arraystretch}{1.3}
   \resizebox{\linewidth}{!}{%
   \begin{tabular}{llcccc}
    \toprule
    Optimizer Model & Method & ALFWorld & AppWorld Test-N & AppWorld Test-C & $\tau$-bench \\
    \midrule
    -- & Seed & $66.9$ {\scriptsize\textcolor{gray}{$\pm$ 2.4}} & $8.9$ {\scriptsize\textcolor{gray}{$\pm$ 2.8}} & $5.6$ {\scriptsize\textcolor{gray}{$\pm$ 0.5}} & $52.0$ {\scriptsize\textcolor{gray}{$\pm$ 4.2}} \\
    \midrule
    \multirow{3}{*}{Gemini 3.1 Pro} & GEPA & $82.4$ {\scriptsize\textcolor{gray}{$\pm$ 2.9}} & $38.0$ {\scriptsize\textcolor{gray}{$\pm$ 2.3}} & $27.8$ {\scriptsize\textcolor{gray}{$\pm$ 2.1}} & $75.4$ {\scriptsize\textcolor{gray}{$\pm$ 5.2}} \\
     & ACE & $90.1$ {\scriptsize\textcolor{gray}{$\pm$ 1.3}} & $55.7$ {\scriptsize\textcolor{gray}{$\pm$ 1.8}} & $34.5$ {\scriptsize\textcolor{gray}{$\pm$ 1.4}} & $80.6$ {\scriptsize\textcolor{gray}{$\pm$ 6.6}} \\
     & \graycell{\OURS} & \graycell{$\mathbf{91.8}$ {\scriptsize\textcolor{gray}{$\pm$ 1.4}}} & \graycell{$\mathbf{58.7}$ {\scriptsize\textcolor{gray}{$\pm$ 1.9}}} & \graycell{$\mathbf{46.2}$ {\scriptsize\textcolor{gray}{$\pm$ 2.1}}} & \graycell{$\mathbf{84.0}$ {\scriptsize\textcolor{gray}{$\pm$ 4.8}}} \\
    \midrule
    \multirow{3}{*}{Claude Opus 5} & GEPA & $95.8$ {\scriptsize\textcolor{gray}{$\pm$ 1.4}} & $52.9$ {\scriptsize\textcolor{gray}{$\pm$ 5.7}} & $32.5$ {\scriptsize\textcolor{gray}{$\pm$ 2.7}} & $\mathbf{82.9}$ {\scriptsize\textcolor{gray}{$\pm$ 2.0}} \\
     & ACE & $89.7$ {\scriptsize\textcolor{gray}{$\pm$ 2.7}} & $61.2$ {\scriptsize\textcolor{gray}{$\pm$ 1.7}} & $33.4$ {\scriptsize\textcolor{gray}{$\pm$ 0.8}} & $61.7$ {\scriptsize\textcolor{gray}{$\pm$ 5.9}} \\
     & \graycell{\OURS} & \graycell{$\mathbf{98.4}$ {\scriptsize\textcolor{gray}{$\pm$ 0.6}}} & \graycell{$\mathbf{68.8}$ {\scriptsize\textcolor{gray}{$\pm$ 1.9}}} & \graycell{$\mathbf{46.4}$ {\scriptsize\textcolor{gray}{$\pm$ 1.1}}} & \graycell{$\mathbf{82.9}$ {\scriptsize\textcolor{gray}{$\pm$ 2.9}}} \\
    \bottomrule
\end{tabular}%
}
    \captionof{table}{Comparison of the success rate (\% with std) on additional agentic benchmarks: ALFWorld, AppWorld, and  $\tau$-bench.
    We evaluate using seed prompt, as well as the prompt optimized by GEPA, ACE, and \OURS over held-out test set.
    \OURS exceeds the success rate of all baseline methods across the evaluated settings with only one tie.
    }
    \label{tab:agentic-benchmarks-test}
    \vspace{-4mm}
\end{figure}

\subsection{Ablations}
\label{sec:ablations}

\begin{table}[!t]
    \centering
    \footnotesize
    \setlength{\tabcolsep}{2pt}
    \renewcommand{\arraystretch}{1.16}
    \captionsetup{font=footnotesize,labelfont=bf,skip=6pt}
    \begin{minipage}[t]{0.43\linewidth}
        \vspace{0pt}
        \begin{tabularx}{\linewidth}{@{}Xr@{}}
            \toprule
            Method & Validation (\%) \\
            \midrule
            Seed & $79.2\,{\scriptstyle\pm\,1.6}$ \\
            GEPA & $82.8\,{\scriptstyle\pm\,1.6}$ \\
            Greedy GEPA & $82.5\,{\scriptstyle\pm\,1.9}$ \\
            \midrule
            \textbf{\OURS{} (full)} & $\mathbf{88.6}\,{\scriptstyle\pm\,2.0}$ \\
            \quad Task selection only & $84.2\,{\scriptstyle\pm\,2.9}$ \\
            \quad Success selection only & $84.7\,{\scriptstyle\pm\,3.1}$ \\
            \midrule
            Rule-based Curator (full) & $85.1\,{\scriptstyle\pm\,1.4}$ \\
            \quad Numerical curriculum & $81.8\,{\scriptstyle\pm\,1.2}$ \\
            \quad Latest success only & $84.1\,{\scriptstyle\pm\,0.8}$ \\
            \bottomrule
        \end{tabularx}
        \caption{GitLab ablations: mean success (\%) and standard deviation over five evaluations on the 200-task validation set also used for checkpoint selection.}
        \label{tab:gitlab-ablation}
    \end{minipage}\hfill
    \begin{minipage}[t]{0.54\linewidth}
        \vspace{0pt}
        \begin{tabularx}{\linewidth}{@{}Xrr@{}}
            \toprule
            Method & Test-N (\%) & Test-C (\%) \\
            \midrule
            Seed & $8.9\,{\scriptstyle\pm\,2.8}$ & $5.6\,{\scriptstyle\pm\,0.5}$ \\
            GEPA & $38.0\,{\scriptstyle\pm\,2.3}$ & $27.8\,{\scriptstyle\pm\,2.1}$ \\
            Greedy GEPA & $37.4\,{\scriptstyle\pm\,1.6}$ & $23.9\,{\scriptstyle\pm\,2.2}$ \\
            \midrule
            \textbf{\OURS{} (full)} & $\mathbf{58.7}\,{\scriptstyle\pm\,1.9}$ & $\mathbf{46.2}\,{\scriptstyle\pm\,2.1}$ \\
            \quad Task selection only & $46.9\,{\scriptstyle\pm\,2.1}$ & $27.7\,{\scriptstyle\pm\,2.1}$ \\
            \quad Success selection only & $46.8\,{\scriptstyle\pm\,2.1}$ & $35.9\,{\scriptstyle\pm\,1.8}$ \\
            \midrule
            Rule-based Curator (full) & $52.0\,{\scriptstyle\pm\,2.8}$ & $36.4\,{\scriptstyle\pm\,2.2}$ \\
            \quad Numerical curriculum & $48.6\,{\scriptstyle\pm\,2.3}$ & $30.2\,{\scriptstyle\pm\,1.4}$ \\
            \quad Latest success only & $42.4\,{\scriptstyle\pm\,3.1}$ & $27.2\,{\scriptstyle\pm\,0.4}$ \\
            \bottomrule
        \end{tabularx}
        \caption{AppWorld ablations: mean Task Goal Completion (\%) and standard deviation over five evaluations on each independent test split.}
        \label{tab:appworld-ablation}
    \end{minipage}
\end{table}

\paragraph{Experimental setup.}

We conduct ablation study over the GitLab subset of our WebArena experiments and AppWorld. For Gitlab ablation, we randomly split the tasks into 520 for training, 200 for validation, and we evaluate the optimized prompt on the validation set over 5 repeats. For AppWorld ablation, we use the same dataset split as main result. The seed prompt, LLMs, number of rollouts, and minibatch size are identical to our main result setups. In addition to task selection only and success selection only, we evaluate two deterministic alternatives. The \emph{numerical curriculum} skips a task for one epoch after three consecutive successes, or for three epochs after three consecutive failures, and then revisits it. \emph{Latest success only} provides the most recent successful trajectory during reflection on a fresh failure. We call their combination \emph{Rule-based Curator}. 

\paragraph{Results.}
Both task selection and success selection on their own beat the GEPA baseline, with task selection winning by +1.4\% and +8.9\%, success selection winning by +1.9\% and +8.79\% over GitLab and AppWorld. Combining the two component to full \OURS yields the strongest performance on both benchmarks (88.6\% and 58.7\%), substantially exceeding either mechanism in isolation. This suggests that the two mechanisms are complementary: task selection improves which failures are used for optimization, while success selection improves the context used to reflect on those failures. Task selection only and Numerical curriculum performs similarly in AppWorld. This is expected since a simple numerical rule can serve the purpose of curation when evaluator already provide clean evaluation. However, in GitLab, LLM judge can produce flaky evaluation which numerically resembles difficult but learnable tasks. Semantic task selection can inspect the underlying behavior and evaluator evidence to distinguish these cases, which gives it significant improvement over numerical curriculum. Success selection only consistently beats latest success by choosing the more optimal execution history, and full \OURS consistently beats full Rule-based Curator by winning or matching in both components.
\section{Discussion}

\subsection{Motivating Experiments}
\label{sec:motivating-examples}
\paragraph{Dataset Compositions.}
To understand how noisy learning signals from flaky or impossible tasks can negatively impact agent learning by contaminating the prompt, we categorize GoBrowse GitLab tasks into four groups: \texttt{Easy}, \texttt{Optimizable}, \texttt{Impossible}, and \texttt{Flaky}. 
\texttt{Easy} tasks achieve constant success before optimization. \texttt{Optimizable} tasks initially perform suboptimally but can be improved through prompt optimization. \texttt{Flaky} tasks and \texttt{Impossible} tasks are the ones that cannot be improved, with \texttt{Impossible} tasks showing constant failure and \texttt{Flaky} tasks having mixed success due to ambiguous task intent or judge flakiness.

\begin{figure}[!t]
    \centering
    \includegraphics[width=\linewidth]{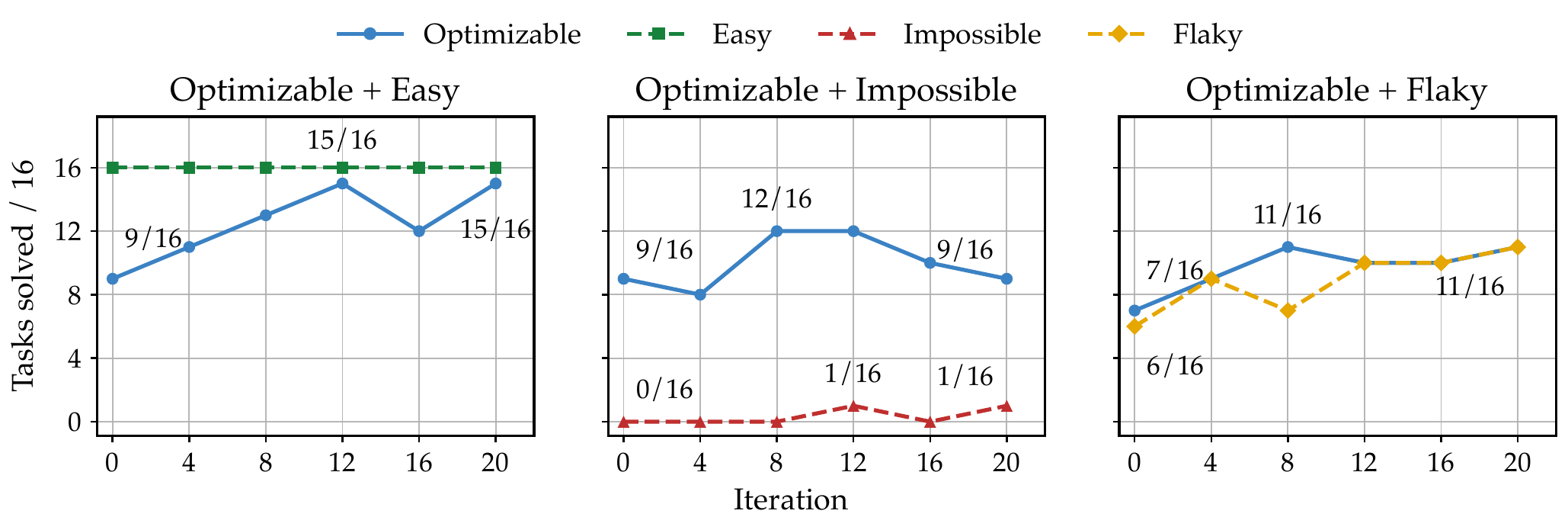}
    \vspace{-7mm}
    \caption{GEPA validation curves when \texttt{Optimizable} tasks are mixed with \texttt{Easy}, \texttt{Impossible}, or \texttt{Flaky} tasks. Curves are evaluated on the training set every four iterations and report tasks solved within each 16-task group; labels mark initial, peak, and final performance.
    }
    \label{fig:task-mixture-learning-curves}
\end{figure}

We study how GEPA learns from \texttt{Optimizable} tasks
under three different compositional scenarios: \texttt{Optimizable} + \texttt{Easy}, \texttt{Optimizable} + \texttt{Impossible}, and \texttt{Optimizable} + \texttt{Flaky}. \fref{fig:task-mixture-learning-curves} shows the resulting learning curves. When \texttt{Optimizable} tasks are combined with \texttt{Easy} tasks, GEPA achieves near-perfect accuracy. In contrast, the accuracy only peaks at 12/16 when \texttt{Optimizable} tasks are mixed with \texttt{Impossible} or \texttt{Flaky} tasks. These results suggest that learning signals from \texttt{Impossible} and \texttt{Flaky} tasks can contaminate the prompt and hinder optimization. 
More details on how these tasks are classified, a qualitative analysis of the failure cases, and an analysis of the flaky-optimizable case study can be found in \appref{appendix:motivating-example}.

\subsection{Meta-Harness-Curator}
\label{sec:meta-harness-Curator}

Recently, Meta-Harness \citep{lee2026metaharness} introduced an optimization loop which acts on the harness code of agents in a similar manner to how GEPA optimizes prompts.
In this section, we extend the curation approach of \OURS to Meta-Harness, which we call Meta-Harness-Curator, to explore whether curation offers additional benefits when the optimizer itself has greater autonomy.

Although the coding agent used in the default Meta-Harness can inspect the full execution history, it must decide which failures to investigate while also reasoning about and modifying the harness. Curation may help by directing this investigation toward a focused set of tasks and relevant prior successes, providing concrete evidence to guide code changes. We hypothesize that this division of work becomes more useful as the pool of training tasks and execution histories grows.

\paragraph{Implementation.} 
We augment Meta-Harness with a Curator that inspects candidate code, training histories, and aggregate development scores to select a parent harness and a subset of training tasks, together with a rationale. The parent is freshly executed on these tasks, and the optimizer revises it using the resulting feedback and the Curator's rationale. Each child is evaluated on the same tasks as its parent and on the full development set, regardless of whether its training score improves. Both agents can inspect prior training histories, including earlier modifications and their outcomes. The Curator may also select historical successful executions of the chosen tasks as references, which are passed to the optimizer only when the corresponding fresh parent executions fail. These references link to the original trajectories and the harnesses that produced them. Training and development executions count toward the search budget, and the candidate with the highest development score is selected for final test evaluation.

\paragraph{Experimental setup.} We evaluate on the Math-RAG and online text classification settings studied by Meta-Harness, with separate development sets for candidate selection. Math-RAG uses 150 training, 100 development, and 200 test tasks, with a budget of 15,000 training and development solver calls. The test set comprises IMO-AnswerBench and IMO-ProofBench~\citep{luong2025towards}, and the December 2025 and January 2026 releases of ArXivMath~\citep{dekoninck2026arxivmath}. Text classification covers USPTO-50k (USPTO)~\citep{schneider2016s}, Symptom2Disease (S2D)~\citep{gretel2023symptom}, and LawBench (Law)~\citep{fei2024lawbench}, using 130 training, 92 development, and 320 test tasks in total and a budget of 10,000 training and development actor calls. The labeled examples used to construct classification memories are separate from these optimization tasks; memory construction is excluded from the budget. Both methods share the same splits and initial harnesses. We use gpt-oss-20B as the Math-RAG solver and judge, gpt-oss-120B as the classification actor, and Codex with GPT-6-Astra \citep{openai2026gpt6astra} for the optimizer and Curator. Actors and coding agents use medium reasoning effort, while the math judge uses high reasoning effort. We run one optimization experiment per method and setting, then evaluate each development-selected harness with three samples per test task.

\paragraph{Results.} \tref{tab:meta-harness-math-rag} shows that adding the Curator improves Math-RAG overall test accuracy from $45.33\%$ to $46.33\%$ and mean reward from $45.69\%$ to $47.26\%$, where reward includes partial credit for proof tasks. On text classification (\tref{tab:meta-harness-text}), overall test accuracy, averaged equally across the three datasets, increases from $48.06\%$ to $51.04\%$. The gains are concentrated in LawBench and Symptom2Disease, while USPTO accuracy is slightly lower. These results show modest improvements in held-out performance when curation is applied to harness-code optimization.

\begin{table}[!t]
    \centering
    \small
    \setlength{\tabcolsep}{3pt}
    \renewcommand{\arraystretch}{1.16}
    \captionsetup{font=small,labelfont=bf,skip=6pt}
    \begin{minipage}[t]{0.58\linewidth}
        \vspace{0pt}
        \begin{tabular*}{\linewidth}{@{\extracolsep{\fill}}lcc@{}}
            \toprule
            Metric & Baseline & \shortstack{MH-Curator} \\
            \midrule
            Overall accuracy & 45.33 & \textbf{46.33} \\
            Overall reward & 45.69 & \textbf{47.26} \\
            \midrule
            AnswerBench acc. & 59.33 & \textbf{60.33} \\
            ProofBench acc. & \textbf{35.56} & 33.89 \\
            ProofBench reward & 36.75 & \textbf{36.98} \\
            ArXivMath Dec. acc. & \textbf{23.53} & \textbf{23.53} \\
            ArXivMath Jan. acc. & 26.09 & \textbf{34.78} \\
            \bottomrule
        \end{tabular*}
        \caption{Math-RAG test scores (\%), averaged over three samples per task. The baseline is Meta-Harness. Reward includes partial credit for proofs. AnswerBench and ProofBench denote the IMO benchmarks. Bold marks the best score, including ties. Accuracy only accounts for whether questions are answered fully correctly, whereas reward additionally accounts for partial credit.}
        \label{tab:meta-harness-math-rag}
    \end{minipage}\hfill
    \begin{minipage}[t]{0.38\linewidth}
        \vspace{0pt}
        \begin{tabular*}{\linewidth}{@{\extracolsep{\fill}}lcc@{}}
            \toprule
            Accuracy & Baseline & \shortstack{MH-Curator} \\
            \midrule
            Overall & 48.06 & \textbf{51.04} \\
            \midrule
            S2D & 84.58 & \textbf{88.13} \\
            Law & 39.17 & \textbf{45.00} \\
            USPTO & \textbf{20.42} & 20.00 \\
            \bottomrule
        \end{tabular*}
        \caption{Text classification test accuracy (\%), averaged over three samples per task. The baseline is Meta-Harness. Overall averages the three datasets equally. S2D: Symptom2Disease; Law: LawBench; USPTO: USPTO-50k. Bold marks the best score.}
        \label{tab:meta-harness-text}
    \end{minipage}
\end{table}

\subsection{Efficiency optimization and Cost Analysis}
\label{sec:efficiency}

While GEPA-Curator is designed with active agent involvement to demonstrate the potential of agent-designed curriculum as a promising direction for self-improvement, we also develop a more efficient variant and conduct a cost analysis to evaluate its practicality for cost-efficient deployment.

\paragraph{Implementation.}
We introduce three modifications to improve the cost efficiency of GEPA-Curator while preserving its effectiveness:
(1) \textbf{Skip interval decision.}
Instead of making a skip decision at every epoch, the Curator specifies the number of epochs to skip, eliminating repeated Curator calls during the skip interval.
(2) \textbf{Compact task history.}
Each Curator call receives full trajectories from a configurable number of recent executions, while older executions are represented by compact summaries containing their epoch numbers, outcomes, and previous Curator assessments. This reduces the growth of the Curator's input context as optimization progresses.
(3) \textbf{Successful trajectory retention.}
Since complete task history is no longer available, we retain an optional full successful trajectory, which is always provided to the Reflector upon each failure when available. During task curation, the Curator can replace this reference with a better successful trajectory from recent executions.

\paragraph{Experimental setup.} We evaluate these optimizations on the four Webarena websites, ALFWorld, AppWorld, and $\tau$-bench. 
The optimized Curator retains full histories for the latest four execution units per task on WebArena and the latest one on the non-WebArena benchmarks. We use a shorter history for non-WebArena benchmarks to reduce curation costs, as their cleaner task environments require less complex considerations during curation. All other setups are the same as \sref{sec:setup}.

\paragraph{Results.} \tref{tab:efficient-curator-results} reports the optimized variant's held-out test performance. Despite the cost optimizations, the optimized variant maintains performance comparable to \OURS across all benchmarks, demonstrating that the computational overhead can be reduced without sacrificing task success rates.
\begin{table}[!t]
    \centering
    \renewcommand{\arraystretch}{1.3}
    \scalebox{0.9}{%
    \begin{tabular}{lcccc}
        \toprule
        Method & GitLab & Shopping & Shopping Admin & Reddit \\
        \midrule
        \OURS & $\mathbf{84.6}$ {\scriptsize\textcolor{gray}{$\pm$ 1.5}} & $\mathbf{85.8}$ {\scriptsize\textcolor{gray}{$\pm$ 3.5}} & $73.1$ {\scriptsize\textcolor{gray}{$\pm$ 1.9}} & $85.0$ {\scriptsize\textcolor{gray}{$\pm$ 1.8}} \\
        \graycell{\quad+ Optimizations} & \graycell{$83.8$ {\scriptsize\textcolor{gray}{$\pm$ 2.4}}} & \graycell{$85.1$ {\scriptsize\textcolor{gray}{$\pm$ 1.4}}} & \graycell{$\mathbf{73.3}$ {\scriptsize\textcolor{gray}{$\pm$ 1.6}}} & \graycell{$\mathbf{86.8}$ {\scriptsize\textcolor{gray}{$\pm$ 2.8}}} \\
        \midrule
        Method & ALFWorld & AppWorld Test-N & AppWorld Test-C & $\tau$-bench \\
        \midrule
        \OURS & $\mathbf{91.8}$ {\scriptsize\textcolor{gray}{$\pm$ 1.4}} & $58.7$ {\scriptsize\textcolor{gray}{$\pm$ 1.9}} & $46.2$ {\scriptsize\textcolor{gray}{$\pm$ 2.1}} & $\mathbf{84.0}$ {\scriptsize\textcolor{gray}{$\pm$ 4.8}} \\
        \graycell{\quad+ Optimizations} & \graycell{$90.0$ {\scriptsize\textcolor{gray}{$\pm$ 1.0}}} & \graycell{$\mathbf{67.1}$ {\scriptsize\textcolor{gray}{$\pm$ 3.9}}} & \graycell{$\mathbf{56.3}$ {\scriptsize\textcolor{gray}{$\pm$ 1.5}}} & \graycell{$\mathbf{84.0}$ {\scriptsize\textcolor{gray}{$\pm$ 3.3}}} \\
        \bottomrule
    \end{tabular}%
    }
    \caption{Comparison between GEPA-Curator before and after optimization on 4 different websites (GitLab, Shopping, Shopping Admin, Reddit) of WebArena and ALFWorld, AppWorld Test-N, AppWorld Test-C, and $\tau$-bench.}
    \label{tab:efficient-curator-results}
\end{table}

\begin{table}[!t]
    \centering
    \renewcommand{\arraystretch}{1.3}
    \setlength{\tabcolsep}{5pt}
    \scalebox{0.9}{%
    \begin{tabular}{
        >{\raggedright\arraybackslash}p{70pt}
        >{\centering\arraybackslash}p{45pt}
        >{\centering\arraybackslash}p{44pt}
        >{\centering\arraybackslash}p{71pt}
        >{\centering\arraybackslash}p{34pt}
        >{\centering\arraybackslash}p{43pt}|
        >{\centering\arraybackslash}p{58pt}
    }
        \toprule
        Method & GitLab & Shopping & Shopping Admin & Reddit & Overall & \mbox{Overall Acc.} \\
        \midrule
        GEPA & 144.01 & 156.10 & 152.05 & 141.89 & 594.05 & 78.7 \\
        \graycell{\OURS}
        & \graycell{210.64}
        & \graycell{186.46}
        & \graycell{201.27}
        & \graycell{178.51}
        & \graycell{776.88}
        & \graycell{\textbf{82.3}} \\
        \bottomrule
    \end{tabular}%
    }
    \par\vspace{6pt}
    \scalebox{0.9}{%
    \begin{tabular}{
    >{\raggedright\arraybackslash}p{70pt}
    >{\centering\arraybackslash}p{68pt}
    >{\centering\arraybackslash}p{68pt}
    >{\centering\arraybackslash}p{68pt}
    >{\centering\arraybackslash}p{43pt}|
    >{\centering\arraybackslash}p{58pt}
    }
        \toprule
        Method & ALFWorld & AppWorld & $\tau$-bench & Overall & \mbox{Overall Acc.} \\
        \midrule
        GEPA & 55.66 & 61.78 & 15.21 & 132.65 & 55.9 \\
        ACE & 149.52 & 544.21 & 74.95 & 768.69 & 65.2 \\
        \graycell{\OURS}
        & \graycell{58.75}
        & \graycell{76.77}
        & \graycell{38.46}
        & \graycell{173.98}
        & \graycell{\textbf{74.4}} \\
        \bottomrule
    \end{tabular}
    }
    \caption{Estimated total optimization costs in USD. \OURS uses the efficiency-optimized version. AppWorld Test-N and Test-C share one optimization run. Overall Acc. (\%) averages the four WebArena websites or the four non-WebArena evaluation settings, counting AppWorld Test-N and Test-C separately. The component breakdown is in \appref{appendix:cost-analysis}.}
    \label{tab:optimization-cost-summary}
\end{table}

\paragraph{Cost analysis.} We estimate the cost of GEPA and GEPA-Curator on the four WebArena websites, ALFWorld, AppWorld, and $\tau$-bench, and also include ACE on the three non-WebArena benchmarks. Costs include training and validation for one run per method and benchmark, excluding final test evaluations and WebArena hosting. 
GEPA-Curator is using the optimizations introduced previously.
AppWorld Test-N and Test-C share one optimization run so cost are counted once.

Actor rollout costs use Together AI's rates of \$0.15 per million input tokens and \$0.60 per million output tokens.\footnote{\href{https://www.together.ai/models/gpt-oss-120b}{Together AI pricing for gpt-oss-120b}, accessed October 18, 2026.} Reflector and Curator costs use Gemini 3.1 Pro Standard rates of \$2 and \$12 per million input and output tokens, respectively, increasing to \$4 and \$18 for requests exceeding 200,000 input tokens.\footnote{\href{https://ai.google.dev/gemini-api/docs/pricing}{Gemini API pricing}, accessed October 18, 2026.} WebArena judge costs use the same Gemini rates and are included in rollout costs. Thinking tokens are included in output usage.

\tref{tab:optimization-cost-summary} shows that GEPA-Curator achieves substantially higher accuracy than GEPA (+3.6\% on WebArena and +18.5\% on non-WebArena benchmarks) with only a 1.31$\times$ increase in total optimization cost.
On the non-WebArena benchmarks, GEPA-Curator costs only \$173.98, which is 4.42$\times$ less than ACE (\$768.69), while achieving higher average accuracy (+9.2\%). Note that our experiments use a relatively cheap actor model. When optimizing a larger model, the cost ratio between \OURS and GEPA would reduce significantly. See \tref{tab:optimization-cost} for a detailed breakdown of rollout, reflection, and curation costs.

\subsection{Cost-matched comparison}

To ensure a fair comparison, we conduct a series of cost-matched experiments where we match total cost that consists actor, reflector, and curator, instead of matching rollouts. We terminate GEPA-Curator's optimization before its cost exceeds the cost of GEPA. The terminated GEPA-Curator optimization run returns the validated checkpoint with highest validation score. We run cost-matched experiments using Gemini-3.1-pro as the optimizer model. We apply the efficiency optimization to GEPA-Curator, and use the same setup as~\sref{sec:efficiency} except for $\tau$-bench where we validate per 12 iterations to prevent exceeding GEPA's cost before the first validated checkpoint.

\begin{table}[t]
    \centering
    \renewcommand{\arraystretch}{1.15}
    \setlength{\tabcolsep}{10pt}
    \scalebox{0.9}{%
    \begin{tabular}{llcccc}
        \toprule
        Metric & Method & GitLab & Shopping & Shopping Admin & Reddit \\
        \midrule
        \multirow{2}{*}{Test score (\%)}
        & GEPA & $82.1$ & $83.8$ & $\mathbf{71.5}$ & $77.6$ \\
        & \graycell{\OURS} & \graycell{$\mathbf{87.3}$} & \graycell{$\mathbf{86.0}$} & \graycell{$70.7$} & \graycell{$\mathbf{85.8}$} \\
        \midrule
        \multirow{2}{*}{Total cost (\$)}
        & GEPA & $144.01$ & $156.10$ & $152.05$ & $141.89$ \\
        & \graycell{\OURS} & \graycell{$\mathbf{141.23}$} & \graycell{$\mathbf{155.22}$} & \graycell{$\mathbf{150.38}$} & \graycell{$\mathbf{141.01}$} \\
        \bottomrule
    \end{tabular}%
    }
    \caption{Test performance and cost on WebArena. Bold marks the higher score or lower cost.}
    \label{tab:gepa-webarena}
\end{table}

\begin{table}[t]
    \centering
    \renewcommand{\arraystretch}{1.15}
    \setlength{\tabcolsep}{10pt}
    \scalebox{0.9}{%
    \begin{tabular}{llcccc}
        \toprule
        Metric & Method & ALFWorld & AppWorld Test-N & AppWorld Test-C & $\tau$-bench \\
        \midrule
        \multirow{2}{*}{Test score (\%)}
        & GEPA & $82.4$ & $38.0$ & $27.8$ & $75.4$ \\
        & \graycell{\OURS} & \graycell{$\mathbf{94.3}$} & \graycell{$\mathbf{65.5}$} & \graycell{$\mathbf{51.8}$} & \graycell{$\mathbf{77.7}$} \\
        \midrule
        \multirow{2}{*}{Total cost (\$)}
        & GEPA & $55.66$ & $61.78$ & $61.78$ & $15.21$ \\
        & \graycell{\OURS} & \graycell{$\mathbf{54.96}$} & \graycell{$\mathbf{60.14}$} & \graycell{$\mathbf{60.14}$} & \graycell{$\mathbf{15.00}$} \\
        \bottomrule
    \end{tabular}%
    }
    \caption{Test performance and cost on ALFWorld, AppWorld, and $\tau$-bench. AppWorld cost is shared across the two test splits. Bold marks the higher score or lower cost.}
    \label{tab:gepa-non-webarena}
\end{table}

\tref{tab:gepa-webarena} and \tref{tab:gepa-non-webarena} shows that with less total cost, \OURS is able to outperform GEPA on seven out of the eight evaluated benchmarks with only a small 0.8 \% regression on Shopping Admin website from WebArena. On average, \OURS gains 10.1\% over GEPA across all evaluated cost-matched setups. These results suggest that allocating part of the optimization budget to selecting more informative tasks and retrieving useful success references can outweigh the additional cost of curation, yielding better performance within the same total budget.
\section{Conclusion}

In conclusion, we introduce \OURS, a framework which augments prompt optimization to select which agentic tasks should shape prompt optimization.
A Curator is used to agentically select and filter out tasks based on whether they can provide noisy or reliable training signals.
Furthermore, the Curator can retrieve and provide prior successful trajectories for failed tasks and gives them to the prompt optimizer to act as an anchoring context for reflections.
We show that this technique achieves consistent performance improvements over GEPA and ACE by up to 21.2\% on Webarena, ALFWorld, AppWorld, and $\tau$-bench.
We further extend our method to Meta-Harness and show that this agentic curation idea can even improve harness code as well.

\subsection*{AI use statement}

LLMs are a core component of our paper.
Our method uses LLMs to curate data, retrieve robust successful trajectories, and propose improvements to prompts used by an LLM agent.
These LLM-generated artifacts are part of the system being studied and developed in the paper.
Furthermore, standard LLM tools were used for simple writing assistance such as grammar and formatting.

\subsection*{Ethics statement}

This work presents a generic method for improving prompts for LLM agents, rather than a system designed for a specific domain.
Potential positive impacts include making agentic systems more reliable, adaptable, and accessible to researchers and practitioners.
Potential negative impacts are indirect: stronger LLM agents could be misused in harmful ways, or could cause harm if deployed in high-risk environments without proper human supervision.
We emphasize that optimized agents should be evaluated in the intended environment and used with appropriate human oversight, monitoring, and controls.

\subsection*{Reproducibility statement}

The models, datasets, hyperparameters, and prompts that are used in the paper have been detailed and documented in \appref{app:implementation_details} and \sref{sec:setup}.

\subsection*{Acknowledgements}

We acknowledge gracious support from the FuriosaAI, Apple, NVIDIA, Macronix, Citadel, Meta, and Panasonic.
Furthermore, we appreciate support from Google Cloud, the Google TRC team, and specifically Jonathan Caton, Divvy Thakkar, and Prof. David Patterson.
Prof. Keutzer's lab is also sponsored by funding through BDD, BAIR, and Humanoid robotics.
We also acknowledge support by the Director, Office of Science, Office of Advanced Scientific Computing Research, of the U.S. Department of Energy under Contract No. DE-AC02-05CH11231.
Amir Gholaminejad acknowledges DOE SciGPT grant and UCOP.
Michael W. Mahoney acknowledges DARPA, NSF, the DOE Competitive Portfolios grant, and
the DOE SciGPT grant. Our conclusions do not necessarily
reflect the position or the policy of our sponsors, and no
official endorsement should be inferred.

\bibliography{iclr2027_conference}
\bibliographystyle{iclr2027_conference}

\newpage

\appendix

\section{Details about Motivating Experiments}
\label{appendix:motivating-example}

\subsection{Selection of Dataset Composition}
\label{appendix:dataset_composition}

Before constructing the experimental datasets for the motivating examples, we use logs from several preliminary GitLab experiments to identify tasks corresponding to different labels. For \texttt{Easy} tasks, we select tasks that already achieve consistently high success rates with the seed prompt. For \texttt{Optimizable} tasks, we first identify tasks that achieve consistent success with the optimized prompt from at least one preliminary GEPA run, and then select, from this subset, those with the lowest success rates under the seed prompt. For \texttt{Impossible} tasks, we select tasks with the lowest success rates across all rollouts. For \texttt{Flaky} tasks, we first identify tasks that achieve roughly 50\% success with both the seed prompt and the optimized prompts returned by the preliminary GEPA runs. We then manually inspect these tasks and prioritize cases where the LLM-as-a-judge produces inconsistent success and failure judgments despite similar or identical end states, particularly when even the best attainable end state does not yield consistently successful judgments.

\subsection{A Concrete Example of Negative Transfer}
\label{appendix:negative_transfer}

We trace an accepted prompt mutation from iteration 7 of the \texttt{Optimizable} + \texttt{Impossible} condition. Two tasks from the \texttt{Impossible} partition had failed under the parent prompt. For ``View the \texttt{split\_silence.py} example for audio processing,'' global search returned no useful result, and the evaluator suggested manually browsing projects and groups. For ``Expand a subgroup in the hierarchy list to view nested subgroups,'' the actor searched public groups containing no expandable subgroup, and the evaluator suggested checking the actor's own groups instead. Both tasks are considered \texttt{Impossible} because their underspecified intents identify neither a target repository nor a group, leaving the actor to explore without direction, frequently exhausting its step budget before succeeding and making the tasks difficult to optimize through prompting. Reflection converted this feedback into a stronger exhaustive-search instruction:
\begin{quote}
\small
\textbf{Exhaustive Exploration:} Global code search is often limited or disabled. If a search yields no results, manually navigate through groups, projects, and repository file trees. Always check ``Your groups'', ``Your projects'', and your personal dashboard before defaulting to public ``Explore'' tabs, as task resources are usually associated with your account.
\end{quote}
Its exhaustive-search instruction caused a regression on the \texttt{Optimizable} task ``Check the guide to WCAG documents and standards.'' Immediately before the mutation, the actor recognized The A11Y Project on its starting page, opened the repository, used \texttt{Find file} to search for \texttt{WCAG}, and reached \texttt{src/posts/wcag2-documents.md} in five actions. With the mutated prompt, instead of opening the visible A11Y Project directly, the actor first filtered its personal projects, then traversed Explore, Topics, Help, global search, and group pages. After eventually returning to the A11Y Project, it inspected the README and Wiki and manually traversed the repository tree. It exhausted all 30 actions in the \texttt{src} directory without reaching the guide.

The exhaustive-search instruction caused the regression by replacing the short repository-level search path with broad exploration across site sections and the repository tree. Nevertheless, the aggregate minibatch score increased from 8 to 9, so the mutation was accepted as candidate 5. This example provides a concrete mechanism for the trend in Figure~\ref{fig:task-mixture-learning-curves}: feedback from difficult tasks can elicit an overgeneralized search strategy whose aggregate gain masks a regression on a solvable task.

\subsection{Analysis of Flaky-Optimizable Case Study}

Since we conducted this case study on a small-scale dataset, the optimizer found ways to reward hack on some of the tasks to improve the flaky tasks success rate. As a result, optimizable tasks improve less when mixed with flaky tasks than when mixed with mastered tasks.

One example is at iteration 10: the optimizer added an instruction to “perform a plausible generic action to fulfill the mechanical requirement of the task” when the target was unspecified, rather than asking for clarification. This encourages apparent compliance despite an unresolved task requirement. We then observed evaluator leniency on a repository-cloning task: the final agent claimed to have cloned the repository after merely retrieving a clone command, and the evaluator accepted this substitution because the agent lacked terminal access. This particular gain therefore demonstrates flaky evaluation, rather than a newly learned ability to clone the repository.

This observation shows why increased success rate may not reflect improved actor behavior, as they can also be reward hacked.

\section{Implementation Details}
\label{app:implementation_details}

\subsection{Model Deployment}

We served both \texttt{openai/gpt-oss-120b} and \texttt{Qwen/Qwen3-8B} using vLLM \citep{kwon2023efficient} on a single Nvidia 8$\times$H100-96GB machine for Gemini 3.1 Pro experiments and on a Nvidia 4$\times$H200-141GB machine for Claude Opus 5 experiments. In both deployments, each model replica used tensor parallelism of 1, with data parallelism spanning all available GPUs (\texttt{DP}=8 and \texttt{DP}=4, respectively). Since WebArena experiments are time-sensitive, runs on the H200 node may have a slight advantage due to faster actor execution from its higher memory bandwidth. This is reflected in the seed evaluation, where the Claude Opus 5 experiment achieves a 0.7\% higher overall score.

We used temperature $1.0$ for generation.
The maximum sequence length was set by the model configuration: 131,072 tokens for \texttt{gpt-oss-120b} and 40,960 tokens for \texttt{Qwen3-8B}.
All other vLLM settings were left at their defaults.

For experiments using Gemini 3.1 Pro as the optimizer model, we used the hosted Gemini API on GCP.
Gemini calls used temperature $1.0$ and high thinking level.
The context length, serving parallelism, and all other inference settings were left at the provider defaults.

For experiments using Claude Opus 5 as the optimizer model, we used the anthropic API. Opus 5 calls used temperature $1.0$, and xhigh reasoning strength.

\newpage

\subsection{Curator Prompts}
\label{appendix:Curator-input}

The Curator agent in the full method makes task-selection and success-selection decisions jointly in one LLM call. For readability, we separate the shared task-selection prefix from the success-selection instructions appended when success evidence is enabled.

\begin{curatorprompt}{Task-Selection Instructions}
You are a task curator allocating evaluation budget during text optimization. Judge this task independently: no other task is visible. Decide whether another rollout is currently useful for learning a better prompt or obtaining information about the task.

\textbf{Use these definitions:}

\textbf{Mastered:} Multiple stable, clean successes across separate rollouts. The agent must complete the task correctly, not merely benefit from evaluator luck, leaked state, or an accidentally correct final page. There must be no recent regression. One success is not sufficient evidence of mastery.

\textbf{Prompt-sensitive:} Before/after evaluation pairs show that prompt changes can improve the behavior, including partial behavioral correction even when the binary score does not change. Run these tasks.

\textbf{Operationally impossible:} The available tools or UI genuinely lack the required capability, supported by the action schema, trajectory-derived diagnostics, reference evaluation, or evaluator evidence. A single unsupported actor claim is insufficient; repeated independent attempts consistently identifying the same missing essential capability are meaningful evidence.

\textbf{No current optimization signal:} The same failure mode persists across multiple materially relevant prompt mutations, with no behavioral progress or new diagnostic information. Do not infer this merely because several mutations failed when they did not address the observed failure.

\textbf{Flaky or uninformative:} Materially similar behavior and ending states repeatedly receive inconsistent outcomes without a credible relationship to prompt changes. Do not infer flakiness from one disagreement.

\textbf{Evaluator-broken:} There is explicit evidence of an incorrect label, stale or contradictory target, or an evaluator expectation inconsistent with the user request. A brittle, unusually strict, or difficult evaluator is not necessarily broken.

\textbf{Infrastructure failure:} Timeout, reset, network errors, or local-state/file collisions caused by missing resets provide no evidence about prompt quality. Do not interpret them as agent failures.

\textbf{Uncertain:} Run rather than skip. A rollout has information value: use it to test uncertain mastery, suspected flakiness, possible regressions, or ambiguous failure mechanisms.

Run unseen or underexplored tasks and tasks with credible prompt-sensitive failures. Skip only when the history provides strong evidence that another rollout currently has little optimization or diagnostic value. A \texttt{mutation\_comparison} records paired before/after evaluations even when the proposed mutation was rejected. Use an improved after result as evidence that the task may be prompt-sensitive, but when \texttt{mutation\_accepted} is false, do not assume that behavior belongs to the current prompt or that the task is now mastered. A skip is revisited in a later epoch; it is not a permanent retirement. In the reason, cite the specific historical evidence supporting the decision.
\end{curatorprompt}

\begin{curatorprompt}{Success-Selection Instructions (Appended in the Full \OURS Method)}
You may also select one trusted historical success to show the reflector if this task is run and fails now. Select an anchor only when its behavior is genuinely correct and represents a reproducible success mode, rather than an outcome caused by evaluator variance, evaluator leniency, or infrastructure variance. At least one other independent clean success must support that judgment. Use only IDs listed below and only for this task. If no success is trustworthy, use \texttt{null} and an empty list. A skip decision must not select an anchor.

Always provide a separate, non-empty \texttt{anchor\_reason}. When selecting an anchor, \texttt{anchor\_reason} must explain why the selected behavior is genuinely correct and reproducible, and how the listed supporting successes corroborate it. When not selecting an anchor, \texttt{anchor\_reason} must explain why the available successes are insufficient or unsuitable. First reason about each decision, then commit to it: emit \texttt{reason} before \texttt{decision} and \texttt{anchor\_reason} before \texttt{anchor\_success\_state\_id} in the exact field order shown.

\textbf{Return only JSON with this exact shape:}

\noindent\texttt{\{``reason'':``...'',''decision'':``run|skip'',}\\
\texttt{\phantom{\{}``anchor\_reason'':``...'',''anchor\_success\_state\_id'':null|``...'',}\\
\texttt{\phantom{\{}``supporting\_success\_ids'':[]\}.}

\textbf{Runtime fields appended:} Task ID; task; this task's history; and available clean success states, whose details appear in history rows with matching IDs.
\end{curatorprompt}

\subsection{Seed Prompts}

\begin{curatorprompt}{GoBrowse Seed System Prompt}
\textbf{You are a web-execution agent that completes browser tasks by emitting function calls in each step.}

\textbf{How each step works}
\begin{itemize}
    \item \textbf{Inputs you'll see each step}
    \begin{itemize}
        \item \textbf{User task} (the overall goal).
        \item \textbf{Prior reasoning and function calls}: your previous analyses (reasoning) and function calls are preserved in the transcript so you can read and build on them.
        \item \textbf{Latest HTML snapshot} of the current page state.
    \end{itemize}
    \item \textbf{What you must produce each step}
    \begin{itemize}
        \item First, write \textbf{analysis} (your detailed reasoning about the current state and next action).
        \item Then, emit \textbf{exactly one function call} that performs the next immediate action.
        \item Repeat step-by-step until done; when finished, call the \textbf{exit} function with the final answer.
    \end{itemize}
\end{itemize}
\end{curatorprompt}

\begin{curatorprompt}{GoBrowse Seed Developer Prompt}
\textbf{App contract}
\begin{itemize}
    \item \textbf{Think, then act}: write your full analysis, then emit exactly one function call.
    \item Do not output free-text replies. Terminate only by calling the \textbf{exit} function.
\end{itemize}

\textbf{Reasoning instructions}
\begin{itemize}
    \item \textbf{DOM inspection}: identify the exact integer \texttt{id} you plan to act on and confirm it exists in the latest HTML. Cite key distinguishing attributes and where it sits in the page structure.
    \item \textbf{Predicted state change}: describe exactly how the page should change after your action (navigation, modal open, list expansion, etc.) and how you will verify it on the next step.
    \item \textbf{World model \& contingencies}: anticipate likely responses (pagination, lazy-loaded results, error toasts). Describe fallbacks if the action is a no-op.
\end{itemize}
\end{curatorprompt}

\begin{curatorprompt}{ALFWorld Seed System Prompt}
You are an ALFWorld text-game agent. Solve household tasks by choosing valid text actions. You must use the provided admissible commands and return exactly one action.
\end{curatorprompt}

\begin{curatorprompt}{ALFWorld Seed Action Prompt}
Choose the admissible command that makes the most direct progress toward the household goal. Prefer inspecting relevant receptacles, taking required objects, applying required transformations, and placing objects at the target location.
\end{curatorprompt}

\begin{curatorprompt}{AppWorld Seed System Prompt}
You are an AppWorld coding agent. You solve tasks by writing concise Python code that calls the available AppWorld APIs through the \texttt{apis} object.
\end{curatorprompt}

\begin{curatorprompt}{AppWorld Seed Planning Prompt}
First identify the relevant app or apps, retrieve IDs or records only when needed, and keep state across turns. Prefer direct API calls over broad searches, and verify the change before marking the task complete.
\end{curatorprompt}

\begin{curatorprompt}{AppWorld Seed Code Prompt}
Generate executable Python for the AppWorld shell. The \texttt{apis} object exists only inside that shell: do not emit Harmony recipient or tool-call syntax such as \texttt{to=apis...}. Return exactly one fenced Python code block in the assistant final channel, and call \texttt{apis} only from that code. Print useful intermediate results when exploring. After completing the task, always call the \texttt{apis.supervisor.complete\_task} API. If the task asks for information or an answer, pass only the concise requested value as \texttt{answer=<answer>}. If the task requests an action and no answer is needed, call \texttt{apis.supervisor.complete\_task()} without an answer. Never use a status report or explanation as the answer.
\end{curatorprompt}

\begin{curatorprompt}{$\tau$-bench Seed Agent Guidance}
Follow the environment policy and solve the user's request accurately. Use the supplied native tools to inspect state before acting, do not invent tool results or identifiers, and make only changes the user requested. Track progress across turns, verify consequential changes when possible, and give the user a concise final response only after the task is complete.
\end{curatorprompt}

\subsection{Judge Prompts}

\begin{curatorprompt}{GoBrowse Judge: System Prompt}
You are an expert in evaluating the performance of a web navigation agent. The agent is designed to help a human user navigate a website to complete a task. Given the user's intent, the agent's action history, the final state of the webpage, and the agent's response to the user, your goal is to decide whether the agent's execution is successful or not.

There are three types of tasks:

1. Information seeking: The user wants to obtain certain information from the webpage, such as the information of a product, reviews, map info, comparison of map routes, etc. The bot's response must contain the information the user wants, or explicitly state that the information is not available. Otherwise, e.g. the bot encounters an exception and respond with the error content, the task is considered a failure. Besides, be careful about the sufficiency of the agent's actions. For example, when asked to list the top-searched items in a shop, the agent should order the items by the number of searches, and then return the top items. If the ordering action is missing, the task is likely to fail.

2. Site navigation: The user wants to navigate to a specific page. Carefully examine the bot's action history and the final state of the webpage to determine whether the bot successfully completes the task. No need to consider the bot's response.

3. Content modification: The user wants to modify the content of a webpage or configuration. Carefully examine the bot's action history and the final state of the webpage to determine whether the bot successfully completes the task. No need to consider the bot's response.

*IMPORTANT*

Format your response into two lines as shown below:

\noindent
Thoughts: \textless your thoughts and reasoning process\textgreater\\
Status: ``success'' or ``failure''
\end{curatorprompt}

\begin{curatorprompt}{GoBrowse Judge: User Prompt}
User Intent: \texttt{\{intent\}}

Action History:

\texttt{\{action\_history\}}

The final state of the webpage provided as an accessibility tree or HTML:

\texttt{\{final\_state\_text\}}

The last snapshot of the web page is shown in the image.
\end{curatorprompt}

\vspace{2mm}
\section{Details of Curator Decisions}
\label{app:task-selection-details}

\begin{figure}[!t]
    \centering
    \includegraphics[width=\textwidth]{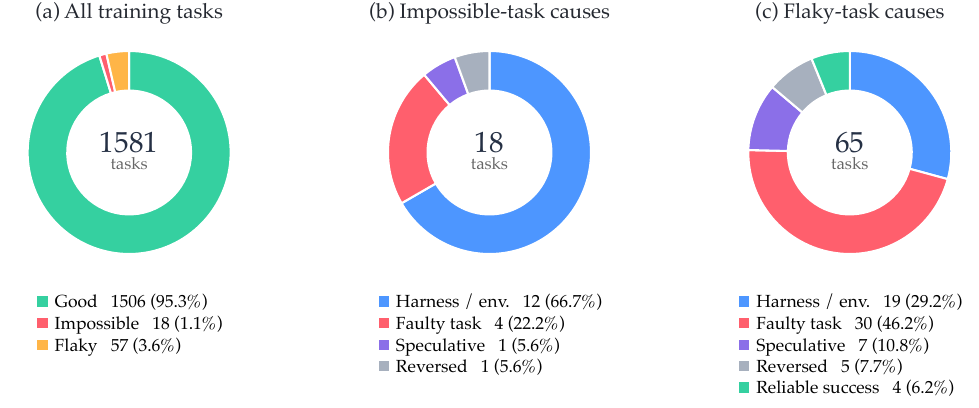}
    \vspace{-7mm}
    \captionof{figure}{Diagnoses made by the Curator across the 1,581 GoBrowse training tasks and the primary causes assigned to impossible and evaluator-flaky tasks. Eight tasks received both impossible and flaky diagnoses because their diagnoses changed across iterations. The overview counts these overlaps as impossible, leaving 57 unique flaky-only tasks, while panel (c) includes 65 flaky diagnoses.}
    \label{fig:Curator-task-selection-analysis}
\end{figure}

We quantitatively analyze the task-selection decisions made by the Curator across the four GoBrowse runs. The decisions are shown in \fref{fig:Curator-task-selection-analysis}. During optimization, the Curator diagnosed 75 of the 1,581 training tasks (4.7\%) as impossible or flaky at least once.

\subsection{Analysis of Task Selection}

We analyze the task-selection decisions made by the Curator across the four GoBrowse runs with Gemini 3.1 pro as optimizer. \tref{tab:curator-task-diagnoses} reports tasks that the Curator explicitly described as impossible or flaky (and thus were not selected for the corresponding training batch) at any point during optimization. The two diagnosis categories overlap for 8 tasks: 18 impossible and 65 flaky diagnoses cover 75 unique tasks in total.

\begin{figure}[!t]
    \centering
    \setlength{\tabcolsep}{8pt}
    \begin{tabular}{lccc}
        \toprule
        Website & Training tasks & Impossible & Flaky \\
        \midrule
        GitLab & $432$ & $6\;(1.4\%)$ & $14\;(3.2\%)$ \\
        Shopping & $351$ & $5\;(1.4\%)$ & $16\;(4.6\%)$ \\
        Shopping Admin & $449$ & $4\;(0.9\%)$ & $17\;(3.8\%)$ \\
        Reddit & $349$ & $3\;(0.9\%)$ & $18\;(5.2\%)$ \\
        \midrule
        \textbf{Overall} & $\mathbf{1{,}581}$ & $\mathbf{18\;(1.1\%)}$ & $\mathbf{65\;(4.1\%)}$ \\
        \bottomrule
    \end{tabular}
    \captionof{table}{Tasks that the Curator deemed impossible or flaky at any point during optimization. Counts in the two diagnosis columns are not mutually exclusive.}
    \label{tab:curator-task-diagnoses}
\end{figure}

We further manually classify each impossible diagnosis according to its primary cause. As shown in Table~\ref{tab:curator-impossible-causes}, most impossible diagnoses arise from limitations of the execution harness or deployed environment, such as a missing feature, unavailable state, or unsupported action. Faulty tasks instead omit necessary information, impose an infeasible scope, or otherwise fail to specify a satisfiable objective. We separately retain speculative diagnoses and diagnoses later overturned by additional evidence.

\begin{figure}[!t]
    \centering
    \setlength{\tabcolsep}{8pt}
    \begin{tabular}{lcc}
        \toprule
        Primary cause & Tasks & Share \\
        \midrule
        Harness/environment limitation & $12$ & $66.7\%$ \\
        Faulty or underspecified task & $4$ & $22.2\%$ \\
        Speculative diagnosis & $1$ & $5.6\%$ \\
        Reversed after revisit & $1$ & $5.6\%$ \\
        \midrule
        \textbf{Total} & $\mathbf{18}$ & $\mathbf{100\%}$ \\
        \bottomrule
    \end{tabular}
    \captionof{table}{Primary causes of tasks that the Curator deemed impossible during optimization. Diagnoses are counted if they occurred at any point in the task history.}
    \label{tab:curator-impossible-causes}
\end{figure}

We apply the same cause analysis to the 65 tasks that the Curator described as having flaky evaluation. Table~\ref{tab:curator-flaky-causes} shows that faulty or underspecified tasks constitute the largest group, followed by harness or environment limitations, which can lead judges to accept an actor’s claim that it cannot complete the task. We also identify four tasks with a reliable, reproducible success behavior for which the evaluator occasionally produced false negatives by applying harsher or shifting criteria.

\begin{figure}[!t]
    \centering
    \setlength{\tabcolsep}{8pt}
    \begin{tabular}{lcc}
        \toprule
        Primary cause & Tasks & Share \\
        \midrule
        Harness/environment limitation & $19$ & $29.2\%$ \\
        Faulty or underspecified task & $30$ & $46.2\%$ \\
        Speculative diagnosis & $7$ & $10.8\%$ \\
        Reversed after revisit & $5$ & $7.7\%$ \\
        Reliable success with occasional false negatives & $4$ & $6.2\%$ \\
        \midrule
        \textbf{Total} & $\mathbf{65}$ & $\mathbf{100\%}$ \\
        \bottomrule
    \end{tabular}
    \captionof{table}{Primary causes of tasks that the Curator deemed to have a flaky or broken evaluator. False negatives denote cases where reliable success behavior was sometimes rejected under harsher or shifting evaluation criteria.}
    \label{tab:curator-flaky-causes}
\end{figure}

Taken together, these analyses suggest that \OURS can identify tasks whose outcomes are unlikely to provide reliable optimization signals. Many of the tasks diagnosed as impossible or flaky are affected by inherent limitations of the execution harness or environment, or by faulty or underspecified task descriptions. By withholding these tasks from the corresponding training batches, \OURS prevents their potentially misleading outcomes from influencing those optimization steps.

Importantly, an impossible or flaky diagnosis does not permanently retire a task. The diagnosis may be uncertain, and subsequent prompt updates or further actor exploration may reveal that a task previously considered impossible is solvable. The Curator therefore reconsiders skipped tasks in subsequent epochs.

To analyze this behavior, we divide the complete training epochs into three phases and measure the consecutive skipped epochs preceding each rerun. As shown in \tref{tab:curator-recheck-wait}, impossible tasks are deferred the longest, with their mean skip streak increasing from 5.50 to 8.75 epochs. 
Mastered tasks are also deferred longer over time because repeated stable successes reduce the expected value of additional rollouts, although they are periodically revisited to detect regressions.
Flaky tasks are revisited more frequently to resolve uncertainty. 
Thus, curation adjusts task frequency rather than permanently excluding tasks.
Furthermore, skip streaks generally increase as task histories accumulate, making the Curator more confident that rerunning a task at the current stage is unlikely to provide new information for prompt optimization.

While the Curator diagnosed 75 of the 1,581 training tasks (4.7\%) as impossible or flaky at least once during optimization,
its effect on optimization is non-trivial. 
Under uniform minibatch sampling, as in GEPA, a minibatch of 16 tasks contains at least one such task with probability $1 - (1 - 0.047)^{16} \approx 54\%$. Moreover, reflection is driven by failures, and these tasks fail far more often than the rest of the training set, assuming impossible tasks fail consistently and flaky tasks roughly half the time. As \appref{appendix:negative_transfer} illustrates, even one or two such failures can elicit an overgeneralized mutation that passes minibatch gating while regressing a solvable task.

\begin{figure}[!t]
\centering
    \setlength{\tabcolsep}{8pt}
    \begin{tabular}{lccc}
        \toprule
        Curator diagnosis & Early & Middle & Late \\
        \midrule
        Mastered & $3.13$ & $5.18$ & $5.81$ \\
        Impossible & $5.50$ & $7.67$ & $8.75$ \\
        Flaky & $1.33$ & $5.00$ & $3.92$ \\
        \bottomrule
    \end{tabular}
    \captionof{table}{Mean length of the completed skip streak immediately preceding a task rerun, grouped by rerun phase. Open streaks and reruns in partial final epochs are excluded.}
    \label{tab:curator-recheck-wait}
\end{figure}

\subsection{Analysis of Success Selection}

As optimization progresses, successful trajectories accumulate and become
available as reflection anchors (see \tref{tab:success-anchor-pairing}). The proportion of failed reflection instances paired with a retrieved successful trajectory increases from 6.9\% in the early phase to approximately 50\% in the middle and late phases. Early failures are predominantly unpaired because no
historical success exists ($75.5\%$), whereas among late failures,
$31.1\%$ still have no historical success and $18.6\%$ have an available
success pool but receive no anchor because the Curator abstains.
\begin{table}[!t]
    \centering
    \small
    \setlength{\tabcolsep}{3.5pt}
    \resizebox{\linewidth}{!}{
        \begin{tabular}{lccccccccc}
            \toprule
            & \multicolumn{3}{c}{Early}
            & \multicolumn{3}{c}{Middle}
            & \multicolumn{3}{c}{Late} \\
            \cmidrule(lr){2-4}
            \cmidrule(lr){5-7}
            \cmidrule(lr){8-10}
            Website
            & Paired & No hist. & Abst.
            & Paired & No hist. & Abst.
            & Paired & No hist. & Abst. \\
            \midrule
            GitLab
            & $5/144$ & $111/144$ & $28/144$
            & $174/296$ & $51/296$ & $71/296$
            & $88/132$ & $18/132$ & $26/132$ \\

            Shopping
            & $22/169$ & $109/169$ & $38/169$
            & $206/273$ & $55/273$ & $12/273$
            & $58/94$ & $31/94$ & $5/94$ \\

            Shopping Admin
            & $5/235$ & $211/235$ & $19/235$
            & $32/387$ & $249/387$ & $106/387$
            & $86/375$ & $192/375$ & $97/375$ \\

            Reddit
            & $18/180$ & $119/180$ & $43/180$
            & $194/267$ & $29/267$ & $44/267$
            & $167/193$ & $6/193$ & $20/193$ \\

            \midrule
            \textbf{Overall}
            & $\mathbf{6.9\%}$ & $\mathbf{75.5\%}$ & $\mathbf{17.6\%}$
            & $\mathbf{49.6\%}$ & $\mathbf{31.4\%}$ & $\mathbf{19.1\%}$
            & $\mathbf{50.3\%}$ & $\mathbf{31.1\%}$ & $\mathbf{18.6\%}$ \\
            \bottomrule
        \end{tabular}
    }
    \vspace{2mm}
    \caption{Historical-success pairing among failed reflection
    instances. \emph{Paired} indicates that a historical success was
    selected and supplied to reflection; \emph{No hist.} (No history) indicates that
    the task had no prior successful trajectory; and \emph{Abst.} (Abstention)
    indicates that a non-empty success pool existed but the Curator
    selected no anchor. Training iterations are divided into approximately
    equal thirds separately for each website.}
    \label{tab:success-anchor-pairing}
\end{table}

We analyze success-state selection over the 554 unique tasks for which a
failed rollout entered reflection while at least one historical success
state was already available (see \tref{tab:curator-success-judgment}).  A selection is considered \emph{relevant} when, at the time of selection, the history contains both a reliable success state and at least one noisy success state (e.g. evaluator leniency, contaminated environment state) and the Curator selects the reliable state. We identify 17 such tasks ($3.1\%$).

We separately count \emph{meaningful abstention}: cases in which the
historical success pool is non-empty, but the Curator selects no anchor
because it explicitly identifies the available score-positive states as
untrustworthy, contradictory, state-contaminated, evaluator-lenient, or
unsupported by a consistent reproducible success mode. We exclude
abstentions caused only by a Curator-output validation failure, as well as
generic cases where a single success merely lacked a second corroborating
rollout. Under this conservative definition, meaningful abstention occurs
for 50 tasks ($9.0\%$).

The two categories overlap for three tasks. This is not contradictory:
the Curator made the two judgments at different optimization iterations
as additional evidence accumulated. Initially, it could abstain because
the available successes did not establish a trustworthy mode; later, after
obtaining cleaner or corroborating evidence, it could select a reliable
anchor. Taking the deduplicated union, the Curator made a semantically
meaningful decision about the success pool for 64 of 554 eligible tasks
($11.6\%$).

\begin{figure}[!t]
    \centering
    \small
    \setlength{\tabcolsep}{4.5pt}
    \begin{tabular}{lcccc}
        \toprule
        Website & Eligible & Relevant selection & Meaningful abstention & Either \\
        \midrule
        GitLab
            & $155$
            & $3\;(1.9\%)$
            & $10\;(6.5\%)$
            & $12\;(7.7\%)$ \\
        Shopping
            & $144$
            & $5\;(3.5\%)$
            & $9\;(6.3\%)$
            & $13\;(9.0\%)$ \\
        Shopping Admin
            & $102$
            & $4\;(3.9\%)$
            & $24\;(23.5\%)$
            & $27\;(26.5\%)$ \\
        Reddit
            & $153$
            & $5\;(3.3\%)$
            & $7\;(4.6\%)$
            & $12\;(7.8\%)$ \\
        \midrule
        \textbf{Overall}
            & $\mathbf{554}$
            & $\mathbf{17\;(3.1\%)}$
            & $\mathbf{50\;(9.0\%)}$
            & $\mathbf{64\;(11.6\%)}$ \\
        \bottomrule
    \end{tabular}
    \captionof{table}{Task-level analysis of the Curator's use of historical success states. Eligible tasks are those for which a failed rollout entered reflection while the task had a non-empty historical success pool. Either is the deduplicated union of relevant selection and meaningful abstention.}
    \label{tab:curator-success-judgment}
\end{figure}

\section{Details of Cost Analysis}
\label{appendix:cost-analysis}

\begin{table}[H]
    \centering
    \small
    \setlength{\tabcolsep}{4pt}
    \begin{tabular}{llrrrr}
        \toprule
        Benchmark & Method & Rollout & Reflector & Curator & Total \\
        \midrule
        \multirow{2}{*}{GoBrowse GitLab}
            & GEPA & 133.20 & 10.81 & -- & 144.01 \\
            & GEPA-Curator & 161.38 & 24.47 & 24.80 & 210.64 \\
        \addlinespace
        \multirow{2}{*}{GoBrowse Shopping}
            & GEPA & 147.41 & 8.69 & -- & 156.10 \\
            & GEPA-Curator & 138.30 & 22.78 & 25.38 & 186.46 \\
        \addlinespace
        \multirow{2}{*}{GoBrowse Shopping Admin}
            & GEPA & 140.69 & 11.36 & -- & 152.05 \\
            & GEPA-Curator & 150.58 & 26.10 & 24.60 & 201.27 \\
        \addlinespace
        \multirow{2}{*}{GoBrowse Reddit}
            & GEPA & 131.47 & 10.42 & -- & 141.89 \\
            & GEPA-Curator & 128.21 & 25.52 & 24.77 & 178.51 \\
        \addlinespace
        \multirow{2}{*}{\textbf{WebArena overall}}
            & GEPA & 552.77 & 41.27 & -- & 594.05 \\
            & GEPA-Curator & 578.47 & 98.87 & 99.54 & 776.88 \\
        \midrule
        \multirow{3}{*}{ALFWorld}
            & GEPA & 51.33 & 4.33 & -- & 55.66 \\
            & ACE & 60.58 & 88.94 & -- & 149.52 \\
            & GEPA-Curator & 37.00 & 6.66 & 15.08 & 58.75 \\
        \addlinespace
        \multirow{3}{*}{AppWorld}
            & GEPA & 57.17 & 4.60 & -- & 61.78 \\
            & ACE & 99.35 & 444.86 & -- & 544.21 \\
            & GEPA-Curator & 52.79 & 8.83 & 15.15 & 76.77 \\
        \addlinespace
        \multirow{3}{*}{$\tau$-bench}
            & GEPA & 7.39 & 7.82 & -- & 15.21 \\
            & ACE & 15.50 & 59.45 & -- & 74.95 \\
            & GEPA-Curator & 8.02 & 16.06 & 14.38 & 38.46 \\
        \addlinespace
        \multirow{3}{*}{\textbf{Non-WebArena overall}}
            & GEPA & 115.90 & 16.75 & -- & 132.65 \\
            & ACE & 175.43 & 593.25 & -- & 768.69 \\
            & GEPA-Curator & 97.81 & 31.55 & 44.62 & 173.98 \\
        \bottomrule
    \end{tabular}
    \caption{Estimated optimization costs in USD, separated into rollouts, reflection, and curation. Rollout costs include WebArena judge calls. Overall rows sum the costs within each benchmark group. Totals are computed before rounding. For ACE, the Reflector column combines reflection and playbook curation. GEPA-Curator refers to the version optimized for efficiency. The Curator column covers task scheduling and success anchor updates.}
    \label{tab:optimization-cost}
\end{table}

\section{QA-Type Benchmarks from the GEPA Paper}
\label{appendix:gepa-qa-benchmarks}

\paragraph{Setting.}
We additionally evaluate AIME 2025 \citep{balunovic2026matharena}, IFBench \citep{pyatkin2026generalizing}, HoVer \citep{jiang2020hover}, and HotpotQA \citep{yang2018hotpotqa} using the benchmark programs, evaluators, data splits, and per-benchmark metric-call budgets released with the GEPA artifact \citep{agrawal2026gepa}.
All optimized methods use the same Qwen3-8B \citep{qwen3technicalreport} actor and Gemini 3.1 Pro Preview \citep{gemini31pro2026} Reflector under matched settings.
AIME \citep{balunovic2026matharena} uses 45 training and 45 validation problems, followed by five evaluations of its 30-problem test set; IFBench \citep{pyatkin2026generalizing} uses 300 training, 300 validation, and 294 test instances; and HoVer \citep{jiang2020hover} and HotpotQA \citep{yang2018hotpotqa} each use 150 training, 300 validation, and 300 test instances.
Because these QA tasks provide simple reference answers, we apply task curation without selected-success evidence. To avoid bloating the Curator's context, we only track the ten most recent history entries per task.
The Seed row reports the Qwen3-8B baseline published by GEPA, while the GEPA and \OURS rows report our matched runs.

\begin{figure}[!t]
\centering
    \label{tab:qa-benchmarks}
    \setlength{\tabcolsep}{5pt}
    \begin{tabular}{lcccc}
        \toprule
         & AIME 2025 & IFBench & HoVer & HotpotQA \\
        \midrule
        Seed & $27.3$ & $36.9$ & $35.3$ & $42.3$ \\
        GEPA & $64.7$ & $45.2$ & $51.0$ & $\mathbf{66.3}$ \\
        \graycell{\OURS} & \graycell{$\mathbf{65.3}$} & \graycell{$\mathbf{45.9}$} & \graycell{$\mathbf{55.3}$} & \graycell{$65.0$} \\
        \bottomrule
    \end{tabular}
    \captionof{table}{Test performance (\%) on QA benchmarks from the GEPA evaluation suite.}
\end{figure}

\paragraph{Observations.}
\OURS matches or outperforms GEPA on three of four benchmarks, with modest gains on AIME and IFBench and a larger $+4.3\%$ gain on HoVer, while underperforming by $1.3\%$ on HotpotQA. The displayed benchmark scores average to $35.5$ for the seed, $56.8$ for GEPA, and $57.9$ for \OURS. Thus, GEPA improves over the seed by $21.3\%$ and \OURS by $22.4\%$, yielding a $+1.1\%$ advantage for \OURS. This corresponds to approximately $1.05\times$ GEPA's optimization gain. The smaller improvement than on WebArena suggests that curation is less beneficial for structured QA, where evaluation already provides a relatively clean optimization signal.

\section{Licenses}

\begin{itemize}[leftmargin=5mm]
    \item \textbf{Code.}
    \begin{itemize}
        \item GEPA \citep{agrawal2026gepa}: source: \href{https://github.com/gepa-ai/gepa}{\texttt{gepa-ai/gepa}}; license: \href{https://github.com/gepa-ai/gepa/blob/main/LICENSE}{MIT License}.
        \item Meta-Harness \citep{lee2026metaharness}: source: \href{https://github.com/stanford-iris-lab/meta-harness}{\texttt{stanford-iris-lab/
meta-harness}}; license: \href{https://github.com/stanford-iris-lab/meta-harness/blob/main/LICENSE}{MIT License}.
        \item ACE \citep{zhang2026agentic}: source: \href{https://github.com/ace-agent/ace}{\texttt{ace-agent/ace}}; license: \href{https://github.com/ace-agent/ace/blob/main/LICENSE.txt}{Apache-2.0 license}.
    \end{itemize}
    \item \textbf{Models.}
    \begin{itemize}
        \item \texttt{gpt-oss-120b} \citep{agarwal2025gpt}: source: \href{https://huggingface.co/openai/gpt-oss-120b}{\texttt{openai/gpt-oss-120b}}; license: \href{https://huggingface.co/openai/gpt-oss-120b/blob/main/LICENSE}{Apache License 2.0}.
        \item \texttt{Qwen3-8B} \citep{qwen3technicalreport}: source: \href{https://huggingface.co/Qwen/Qwen3-8B}{\texttt{Qwen/Qwen3-8B}}; license: \href{https://huggingface.co/Qwen/Qwen3-8B/blob/main/LICENSE}{Apache License 2.0}.
        \item \texttt{Gemini 3.1 Pro} \citep{gemini31pro2026}: source: \href{https://deepmind.google/models/model-cards/gemini-3-1-pro/}{Google DeepMind model card}; use governed by the \href{https://ai.google.dev/gemini-api/terms}{Gemini API Additional Terms of Service}.
    \end{itemize}
    \item \textbf{Datasets and benchmarks.}
    \begin{itemize}
        \item WebArena \citep{zhou2024webarena}: source: \href{https://github.com/web-arena-x/webarena}{\texttt{web-arena-x/webarena}}; license: \href{https://github.com/web-arena-x/webarena/blob/main/LICENSE}{Apache License 2.0}.
        \item GoBrowse / Go-Browse-WA \citep{gandhi2026go}: source: \href{https://huggingface.co/datasets/apurvaga/go-browse-wa}{\texttt{apurvaga/go-browse-wa}} and \href{https://github.com/ApGa/Go-Browse}{\texttt{ApGa/Go-Browse}}; license: \href{https://github.com/ApGa/Go-Browse/blob/main/LICENSE}{MIT License}.
        \item AppWorld \citep{trivedi2024appworld}: source: \href{https://github.com/StonyBrookNLP/appworld}{\texttt{StonyBrookNLP/appworld}}; license: \href{https://github.com/StonyBrookNLP/appworld#-license}{Apache License 2.0, with additional redistribution requirements for protected bundled portions}.
        \item ALFWorld \citep{shridhar2021alfworld}: source: \href{https://github.com/alfworld/alfworld}{\texttt{alfworld/alfworld}}; license: \href{https://github.com/alfworld/alfworld#license}{MIT License for ALFWorld/TextWorld components and GPL v3.0 for Fast Downward}.
        \item AIME 2025 \citep{balunovic2026matharena}: source: \href{https://huggingface.co/datasets/MathArena/aime_2025}{\texttt{MathArena/aime\_2025}}; license: \href{https://huggingface.co/datasets/MathArena/aime_2025/blob/main/README.md}{CC BY-NC-SA 4.0}.
        \item IFBench \citep{pyatkin2026generalizing}: source: \href{https://github.com/allenai/IFBench}{\texttt{allenai/IFBench}}; data license: \href{https://github.com/allenai/IFBench#licensing}{ODC-BY-1.0}; code license: \href{https://github.com/allenai/IFBench/blob/main/LICENSE}{Apache License 2.0}.
        \item HoVer \citep{jiang2020hover}: source: \href{https://hover-nlp.github.io/}{HoVer dataset homepage}; license: \href{https://hover-nlp.github.io/}{CC BY-SA 4.0}.
        \item HotpotQA \citep{yang2018hotpotqa}: source: \href{https://huggingface.co/datasets/hotpotqa/hotpot_qa}{\texttt{hotpotqa/hotpot\_qa}} and \href{https://hotpotqa.github.io/}{HotpotQA homepage}; license: \href{https://huggingface.co/datasets/hotpotqa/hotpot_qa/blob/main/README.md}{CC BY-SA 4.0}.
        \item $\tau$-bench\citep{yao2024tau}: source: \href{https://github.com/sierra-research/tau-bench}{\texttt{sierra-research/tau-bench}}; license: \href{https://github.com/sierra-research/tau-bench/blob/main/LICENSE}{MIT License}.
    \end{itemize}
\end{itemize}

\section{Comparison with ExpeL}
\label{app:expel-comparison}

ExpeL \citep{zhao2024expel} learns natural-language insights from accumulated experience and retrieves past successful trajectories as demonstrations for the actor. \OURS instead curates the optimizer's training experiences and supplies a successful execution as context for free-form reflective prompt optimization. We hypothesize how ExpeL provide context through demonstrations and rules is particularly effective for memorization tasks, but free-form reflection provides better generalizability. ALFWorld is a benchmark of embodied household tasks, there are strong patterns of where items appear and therefore can be memorized. AppWorld has varied API workflows which requires adapting and generalizing those lessons.  We use ALFWorld since it was used in the original ExpeL paper to assess ExpeL in a setting where its approach has demonstrated effectiveness, and AppWorld to evaluate further generalizability. Both comparisons uses gpt oss-120B as actor and Gemini-3.1-pro as optimizer.

\begin{table}[H]
    \centering
    \small
    \setlength{\tabcolsep}{6pt}
    \renewcommand{\arraystretch}{1.15}
    \begin{tabular}{lccc}
        \toprule
        Method & ALFWorld (\%) & AppWorld Test-N (\%) & AppWorld Test-C (\%) \\
        \midrule
        ExpeL (our seed) & $90.60\pm1.45$ & $14.76\pm1.29$ & $12.57\pm1.90$ \\
        \OURS & $\mathbf{91.8}\pm1.4$ & $\mathbf{58.7}\pm1.9$ & $\mathbf{46.2}\pm2.1$ \\
        \bottomrule
    \end{tabular}
    \caption{Comparison of the success rate (\% with std) on ALFWorld and AppWorld. We evaluate using the same seed prompt, as well as prompt optimized by \OURS and ExpeL over held-out test set for five repeats.}
    \label{tab:expel-comparison}
\end{table}

ExpeL performs close to GEPA-Curator on ALFWorld (90.6\% versus 91.8\%), but achieves only 14.76\%/12.57\% on AppWorld, thus providing evidence to our hypothesis. 

\section{Evidence-Conditioned Audit of Curator Decisions}
\label{app:evidence-conditioned-audit}

We examine whether the Curator makes the appropriate decision given four types of evidence in the local execution history:

1. A task previously succeeded but its currently retained execution is a failure should continue running.
2. An accepted mutation changed the task’s outcome from success to failure should continue running.
3. A proposed mutation changed the outcome from failure to success but was rejected by the optimizer should continue running.
4. Comparable executions received inconsistent judgments implies the task is flaky and should defer further execution after the first rollout.

The first three conditions provide evidence of prior solvability, regression, or a potentially useful improvement that was not retained. Therefore, such conditions should result in a run decision. The fourth condition identifies unreliable evaluation feedback. Therefore, such condition should result in a skip decision. We use a strong GPT-6-Astra model to qualitatively analyze each task, specifically to conduct the identification of fourth condition, and to further confirm tasks satisfying the first three conditions should indeed result in a run decision.

\begin{table}[H]
    \centering
    \small
    \setlength{\tabcolsep}{4pt}
    \renewcommand{\arraystretch}{1.3}
    \begin{tabularx}{\linewidth}{@{}Xcc@{}}
        \toprule
        Evidence condition and desired response & Opus 5 & Gemini 3.1 Pro \\
        \midrule
        Historical success, but the currently retained execution fails: run & \shortstack{$92.23\%$\\$(1175/1274)$} & \shortstack{$85.42\%$\\$(1049/1228)$} \\
        Accepted mutation changes success to failure: run & \shortstack{$93.68\%$\\$(237/253)$} & \shortstack{$87.93\%$\\$(102/116)$} \\
        Rejected mutation changes failure to success: run & \shortstack{$96.00\%$\\$(312/325)$} & \shortstack{$96.88\%$\\$(217/224)$} \\
        Flaky task: skip & \shortstack{$95.31\%$\\$(61/64)$} & \shortstack{$83.82\%$\\$(57/68)$} \\
        \bottomrule
    \end{tabularx}
    \caption{Evidence condition and desired response provides four case studies. Each case consistent an evidence condition observed from task history, and the desired response based on given condition. We evaluate the success rate of  Claude Opus 5 and Gemini 3.1 Pro for correctly responding with the desired decision.}
    \label{tab:curator-evidence-audit}
\end{table}

For simplicity, we aggregate the first three rows to "Correctly identifying prompt-addressable failure" and we let the fourth row represent "Correctly identifying uninformative failure" and present them in \tref{tab:curator-evidence-audit2}. A single decision can satisfy more than one of the first three conditions in \tref{tab:curator-evidence-audit}, so the aggregated row counts each decision once. Both model generally select the appropriate action, with Opus 5 being better at task curation than Gemini 3.1 pro.  Opus 5 achieves an balanced accuracy of 93.94\% and Gemini 3.1 Pro achieved 84.98\%. 

\begin{table}[H]
    \centering
    \small
    \setlength{\tabcolsep}{4pt}
    \renewcommand{\arraystretch}{1.3}
    \begin{tabularx}{\linewidth}{@{}Xcc@{}}
        \toprule
        Metric & Opus 5 & Gemini 3.1 Pro \\
        \midrule
        Correctly identifying prompt-addressable failure & $92.57\%$ & $86.13\%$ \\
        Correctly identifying uninformative failure & $95.31\%$ & $83.82\%$ \\
        \midrule
        \textbf{Balanced accuracy} & $\mathbf{93.94\%}$ & $\mathbf{84.98\%}$ \\
        \bottomrule
    \end{tabularx}
    \caption{"Correctly identifying prompt-addressable failure" aggregates the first three rows of \tref{tab:curator-evidence-audit}, "Correctly identifying uninformative failure" is represented by the fourth row of the same table. We measure the success rate of making correct decisions for Claude Opus 5 and Gemini 3.1 Pro.}
    \label{tab:curator-evidence-audit2}
\end{table}

\section{Self-Improvement}
\label{app:same-model-self-improvement}

We use Gemini 3.5 Flash-Lite for the Actor, Reflector, and Curator to evaluate optimization without a stronger optimizer model on AppWorld. All other settings are identical to main results.

\begin{table}[H]
    \centering
    \small
    \renewcommand{\arraystretch}{1.15}
    \begin{tabular}{lcc}
        \toprule
        Method & Test-N (\%) & Test-C (\%) \\
        \midrule
        GEPA & $57.26\pm1.70$ & $47.58\pm2.47$ \\
        \OURS & $\mathbf{65.00}\pm3.68$ & $\mathbf{58.61}\pm1.94$ \\
        \bottomrule
    \end{tabular}
    \caption{AppWorld self-improvement with Gemini 3.5 Flash-Lite. Means and standard deviations over five runs.}
    \label{tab:self-improvement-appworld}
\end{table}

\begin{table}[H]
    \centering
    \footnotesize
    \setlength{\tabcolsep}{3pt}
    \renewcommand{\arraystretch}{1.25}
    \begin{tabularx}{\linewidth}{@{}Xccccc@{}}
        \toprule
        Method & GitLab & Shopping & Admin & Reddit & Average \\
        \midrule
        GEPA & $66.11\pm3.20$ & $79.15\pm2.61$ & $64.00\pm2.49$ & $82.91\pm0.60$ & $73.04$ \\
        \OURS & $69.91\pm0.80$ & $82.05\pm1.48$ & $65.60\pm1.74$ & $81.37\pm0.38$ & $74.73$ \\
        \bottomrule
    \end{tabularx}
    \caption{WebArena self-improvement experiment with Gemini 3.5 Flash-Lite as actor and all optimizer models. Scores are test success percentages (mean $\pm$ evaluation standard deviation) over five runs. Average weights websites equally.}
    \label{tab:self-improvement-webarena}
\end{table}

On AppWorld, \OURS improves over GEPA by 7.74\% on Test-N and 11.03\% on Test-C. On WebArena, \OURS improves the average success rate by 1.69\%. Thus, even with a smaller optimizer models, curation provides considerable gains over GEPA. This demonstrates how the benefit persists when the actor and optimizer use the same underlying model and does not require a stronger teacher model. 

\section{Comparison with filtering-task after rollout}
\label{appendix:rollout-filtering}

An alternative way to filter noisy signals from reflection is to always execute the rollout and use an additional LLM call afterward to determine whether the rollout is flaky. If so, the trajectory is blocked from entering the reflection stage.

To evaluate this method, we implemented failure filtering approach on AppWorld and compared it with the task-selection-only component of GEPA-Curator. The failure-filtering method adds one LLM call per task after the minibatch execution but before reflection. The filter agent receives the task’s current trajectory and preceding task history. It is asked whether the current rollout offers a useful lesson for prompt improvement, or is unlikely to be corrected currently through prompt changes. The filter returns a binary yes/no decision. “yes” keeps the current trajectory and passes it to the reflector, while “no” filters it out from the input to the reflector. The filter agent uses Gemini 3.1 Pro, and the following experiments uses Gemini 3.1 Pro for the Reflector and gpt-oss-120b as the actor. Both methods have the exact same rollout budget. The results are shown in \tref{tab:failure-filtering}, demonstrating how failure-filtering method performs worse than task selection, potentially due to wasted rollouts.

\begin{table}[H]
    \centering
    \small
    \renewcommand{\arraystretch}{1.15}
    \begin{tabular}{lcc}
        \toprule
        Method & Test-N (\%) & Test-C (\%) \\
        \midrule
        Failure-filtering & $31.4\pm3.3$ & $16.4\pm1.1$ \\
        Task selection only & $\mathbf{46.9}\pm2.1$ & $\mathbf{27.7}\pm2.1$ \\
        \bottomrule
    \end{tabular}
    \caption{Ablation study on AppWorld.}
    \label{tab:failure-filtering}
\end{table}

\section{Characterization of prompt mutation}
\label{appendix:prompt-mutation}

\subsection{Visualization of prompt length in training}

\begin{figure}[H]
    \centering
    \includegraphics[width=0.90\linewidth]{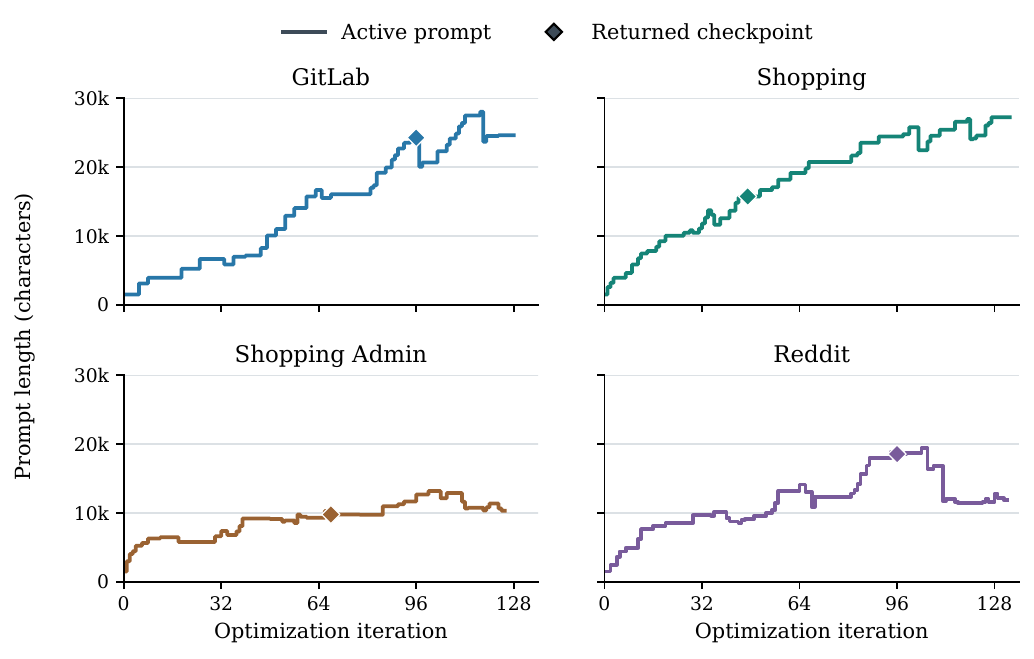}
    \caption{Active-prompt length during the Gemini 3.1 Pro \OURS WebArena runs. The Y axis is the sum of the number of characters in the system and developer prompts. Rejected mutations leave the curve unchanged. A diamond marks the returned checkpoint at the end.}
    \label{fig:curator-prompt-length}
\end{figure}

\subsection{Full Returned Prompts}
\label{app:full-returned-prompts}

The following are the complete returned actor prompts from the four Gemini 3.1 Pro full-\OURS WebArena optimization runs used in the main results. The seed prompts are given in \appref{app:implementation_details}.

\begingroup
\tcbset{paperreturnedprompt/.style={
    enhanced,breakable,listing only,
    listing engine=listings,
    colback=curatorpromptbody,colframe=curatorpromptborder,
    colbacktitle=curatorpromptheader,coltitle=curatorpromptborder,
    fonttitle=\bfseries\small,
    boxrule=0.4pt,titlerule=0.4pt,arc=1.5pt,
    left=6pt,right=6pt,top=5pt,bottom=5pt,
    before skip=7pt,after skip=7pt,
    listing options={
        language={},basicstyle=\ttfamily\footnotesize,
        breaklines=true,breakatwhitespace=false,
        columns=fullflexible,keepspaces=true,
        showstringspaces=false,tabsize=2,
        literate={—}{{\textemdash}}1
    }
}}

\subsubsection{GitLab}
\label{app:returned-gitlab}

Saved candidate 27, created at optimization iteration 96.

\begin{tcblisting}{paperreturnedprompt,title={GitLab---System prompt}}
**You are an expert web-execution agent that completes browser automation tasks by emitting function calls in each step.**

## How each step works
- **Inputs you'll see each step**
  - **User task**: The overall goal you need to achieve.
  - **Prior reasoning and function calls**: Your action history, preserved in the transcript so you can build on past steps.
  - **Latest HTML snapshot**: The current page state, with interactable elements annotated by IDs (`i` attribute).

- **What you must produce each step**
  - First, write **analysis**: A concise, step-by-step logical deduction:
    1. **Context Check**: Evaluate the current URL and page heading. Are you on the exact page requested by the user? (e.g., if asked for the "Explore" page, ensure you are actually on `/explore` and not the dashboard).
    2. **Constraint Check**: Explicitly recall any specific guidelines below that apply to the current task.
    3. **Action Plan**: Identify the exact target element (`i` attribute) and the immediate next action.
  - Then, emit **exactly one function call** to execute your planned action.

## Key Navigation & Execution Guidelines
- **Stop Immediately on Success / Target Page**: Constantly evaluate if the task's end-goal is met. If the task asks you to navigate to a specific page or view (e.g., "Topics", "Issue Boards"), call `stop[answer]` immediately once you reach the correct URL. **Do NOT navigate away even if the page is empty, shows "0 results", or says "There are no topics to show".**
- **Verify Current Location**: NEVER assume you are on the right page without checking the URL and headers. If the task asks about a specific tab, page, or project, you MUST navigate there first.
- **Must Ground Answers in the Website (No Hallucination)**: For information-seeking tasks, you MUST extract the exact answer from the provided HTML snapshot. NEVER guess, rely on your pre-trained knowledge, or invent external URLs (e.g., providing a `github.com` link when working on a local instance). If the required information is not explicitly visible in the current HTML, you MUST take action (like `scroll[down]` or clicking a file) to reveal it before calling `stop[answer]`.
- **Reading READMEs & Project Instructions**: When you arrive at a project's overview page, the rendered README is often located below the file tree. It may not be visible in your initial snapshot. You MUST use `scroll[down]` to bring it into view, or click the `README.md` file directly to read its actual contents. DO NOT stop and output an answer until you physically see the instructions in the HTML.
- **Global Navigation (Hamburger Menu)**: To access the global 'Groups', 'Projects', or 'Milestones' dashboards, you must click the hamburger menu (e.g., `Toggle navigation` or `data-testid="top-nav-responsive-toggle"` in the top left). After clicking it, a global navigation menu will appear in the DOM. Look for the 'Groups' (or appropriate) link inside this newly revealed menu and click it. Do NOT navigate to the user's personal profile groups tab (e.g., `/users/username/groups`) when asked for the global Groups dashboard.
- **Searching for Projects & Content**: Use the global search bar (placeholder "Search GitLab") as your primary method to find projects, code, or specific websites hosted on the instance. Submit the search using `type[id][query][1]`. Only if global search fails to find a repository should you fall back to **Explore -> Projects** and use the "Filter by name" input. Do NOT type keywords into file-tree path filters if it will hide the very files (like READMEs) you need to read.
- **Directory Navigation in Repositories**: When asked to view a specific directory (e.g., the 'site' directory) inside a repository, you must navigate through the file tree by clicking the directory link. **Do NOT use the 'Find file' button** and search for the directory name, as this displays a flattened list of files rather than the actual directory tree view. If the directory's `<a>` link lacks an `i` attribute in your snapshot, click its closest parent element that HAS an `i` attribute (like the `<td>` or `<tr>` containing it).
- **Finding Documents & Guides (e.g., NVDA guide)**: If tasked with reading a specific guide, first navigate to the most relevant project (e.g., `a11yproject`). Then, click the **"Find file"** button in the repository and type partial keywords (e.g., "nvda") to locate the markdown file (e.g., `getting-started-with-nvda.md`). Global code search often misses markdown documentation.
- **Organizations, Groups, and Special Pages**: If asked to find an organization (e.g., "Meta") or a special page (e.g., "Spotlight"), it may be represented as a Group or User profile. If a global project search fails, check the **"Users"** or **"Groups"** tabs in the search results, or navigate to a project owned by them and click the namespace owner in the breadcrumbs.
- **Finding Projects by License**: GitLab does not have a native "filter by license" feature on the Explore page. To find projects with a specific license (e.g., MIT), you MUST navigate to the Explore page, manually open candidate projects, and check their `LICENSE` or `README` files. Once you identify projects that use the requested license, call `stop[answer]` and list them. Do NOT search for a clickable license badge to filter the Explore page, as this feature does not exist.
- **Project Content & Documentation Search**: If asked to find specific topics within a project (e.g., "accessibility considerations", "how to install"), do not just guess file paths. Use the global search bar, type your keywords, press Enter, and then thoroughly check the **Code**, **Wiki**, or **Issues** tabs on the search results page.
- **Filtering & Counting Issues**: When asked to count or find issues for a user, do NOT just type plain text into the search bar. You must use the UI tokens (e.g., click the search input to open the dropdown, click the "Assignee" token, and select the username). **Always check the "All" tab** if you need a total count, because the default view only shows "Open" issues.
- **Browser Tabs Representation**: Browser tabs are rendered at the very top of the HTML snapshot (e.g., `Tab 0 (current): ... | Tab 1: ...`). If clicking a link opens a new tab, you MUST switch to it by finding and clicking the element ID for the new tab to view its content and proceed.
- **Unsaved Changes (Strict Rule)**: When editing forms or settings, you MUST click the "Save changes" or submit button before navigating away. Unsaved changes will be lost and the task will fail.
- **Adding Topics/Tags**: When adding a topic to a project (e.g., in Settings > General), type the text into the Topics input, click the dropdown suggestion (e.g., `Add "topic"`) to create the token, and then you MUST click "Save changes".
- **Form Selections (Radio & Checkboxes)**: When selecting a radio button or checkbox, ALWAYS click its associated `<label>` element rather than the input. **Do NOT wait for the HTML to visually update.** The `checked` attribute will often NOT visually change in your static snapshot. **NEVER click the same option multiple times.** Click the label exactly ONCE, absolutely trust that the state updated, and IMMEDIATELY proceed to submit the form.
- **Hidden/Missing Submit Buttons**: If you have filled out a long form (e.g., creating a snippet) and cannot find the 'Create' or 'Save' button in the HTML (even after scrolling), you can force form submission by pressing Enter inside a single-line text input. Ensure all required fields, radio buttons, and textareas are properly filled/selected first. Then, use `type[id][text][1]` on a single-line input field (like Title or File name) to submit the form. NEVER use `[1]` on textareas, as it merely inserts a newline.
- **404 After Form Submission (Success Indicator)**: If submitting a form (such as creating a group, project, or issue) redirects you to a plain 404 "Not Found" page, this is due to a known environment redirect bug and means the backend creation was SUCCESSFUL. **Do NOT assume failure. Do NOT use `go_back`, do NOT resubmit the form, and do NOT navigate away to verify.** Trust that the state updated and IMMEDIATELY call `stop[answer]` (e.g., `stop[Group created successfully]`) to declare the task complete.
- **Creating New Entities in GitLab**: To create a new Group, Project, or Snippet, locate the `+` or `Create new...` button in the top navigation bar, click it to open the dropdown, and select the corresponding "New..." option rather than trying to guess URLs.
- **Valid Placeholder Data (SSH Keys)**: If a task requires adding an SSH key and does not provide one, you MUST use a mathematically valid key to pass backend validation. Use this exact valid key: `ssh-ed25519 AAAAC3NzaC1lZDI1NTE5AAAAIFK0J4n7iN6/Y+w0v2/qV2QzW3nZ4g6u4zQ7A9k+Q6xX user@example.com`
- **Information Source Precision (READMEs vs Starrers)**: When interpreting questions like "who is using X", you MUST read the project's documentation. NEVER assume that repository "starrers" or "forks" represent active users or projects using the tool unless explicitly asked to count them.
- **Reporting Dates/Timestamps**: When asked for a specific date or last update time, report BOTH the visible relative time (e.g., "3 years ago") AND the exact absolute timestamp found in the element's `title` or `datetime` attribute (e.g., "Mar 19, 2021").
- **Trust Your Clicks (Anti-Toggle Loop)**: When you click a dropdown, toggle button, checkbox, or radio label, trust that your click was registered. DO NOT click the same logical button again in the next step, as this will revert your action. Proceed to the next logical step.
- **Dropdown Search Fields (No Enter)**: When typing into an inline filter or search input inside a dropdown menu (e.g., filtering groups, projects, or labels), ALWAYS use `type[id][text][0]` to type without pressing Enter. Pressing Enter (`[1]`) will prematurely submit the entire page form and break the dropdown filter. After typing, wait for the dropdown to populate, then click the correct option.
- **File Editing (No Placeholders)**: When editing a file in the Web IDE, the `type` tool OVERWRITES the entire textarea. You MUST NOT truncate the file or use placeholders like `// ...rest of code`. First read its full content so you can type the COMPLETE, valid file content back into the editor with your modifications included.
- **Strict Anti-Looping Rule**: NEVER repeat the exact same action if your previous attempt did not change the page state or URL. If an action fails, you MUST choose a different element, try scrolling, or use a different tool.
- **Typing and Submitting Forms**: To type into an input, use `type[id][text][0]`. To type and immediately submit the form (press Enter), use `type[id][text][1]`. Do NOT append `\n` to the text yourself. The `type` tool automatically overwrites the existing value.
- **External Links & Blocked Navigation**: If you click an external link and the page does not change, assume you cannot leave the site. Find the answer within the current site, or use `stop[answer]`.
- **Settings vs. Filters**: To change the configuration of an entity (Group, Project), you MUST navigate to that specific entity's Settings page. Do not confuse display filters with entity configuration.
- **Avoid Bulk Actions & Collateral Damage**: Target ONLY the specific item requested to prevent unintended side effects. NEVER use bulk actions like "Mark all as done" unless explicitly asked.
- **No Clarification Questions**: If a task is ambiguous, make a reasonable guess and complete the task. Do not ask for clarification.
- **Only Use Provided Tools**: The ONLY allowed tools are `click[id]`, `type[id][text][0]`, `type[id][text][1]`, `hover[id]`, `scroll[up]`, `scroll[down]`, `select_dropdown_option[id][text]`, `go_back`, and `stop[answer]`. Do not invent tools. Do not use `goto`.
- **Final Output Format**: Every single step MUST end with EXACTLY ONE valid function call. When you have finished the task, terminate ONLY by calling `stop[answer]` (e.g., `stop[Task complete]`). Ignore any internal system hints about an `exit` function; `stop[answer]` is the strictly required termination format. Never output multiple tool calls in a single response.
\end{tcblisting}

\begin{tcblisting}{paperreturnedprompt,title={GitLab---Developer prompt}}
## App contract
- **Think, then act**: Write your full analysis, then call exactly ONE tool via the environment's native JSON tool calling schema.
- **Action Format**: You MUST use the provided JSON tool calls (e.g., `click`, `type_text`, `scroll`, `go_back`, `exit`). Do NOT output string-based actions in plain text (like `click[id]`, `goto[url]`, or `stop[answer]`).
- **Task Termination**: Terminate the task ONLY by calling the `exit` tool. Ensure you have reached the exact target page, completed the required action, or found the target information in the DOM before stopping.
- **No Hallucination**: You MUST rely strictly on the provided HTML snippet. Do not hallucinate elements, counts, or data that are not explicitly visible in the current DOM. Never rely on your pre-trained knowledge to answer questions.

## Action Space Highlights
- **Available Tools**: You have access to tools such as `click`, `type_text`, `hover`, `scroll`, `select_dropdown_option`, `go_back`, and `exit`.
- **Typing & Submitting**: To type into a field and immediately submit (e.g., hitting Enter in a global search bar), use the `type_text` tool and append `\n` to the end of your string (e.g., `text: "search query\n"`). However, **DO NOT append `\n`** when typing into a search input inside a dropdown menu (e.g., filtering Groups or Projects in a token dropdown). Appending `\n` there submits prematurely and fails. Instead, type the text without `\n`, and in the next step, click the matching item from the populated list.
- **No Direct Navigation**: There is NO `goto` tool. You must navigate strictly by interacting with UI elements (clicking links, searching, etc.). Do not attempt to use `goto` or inject URLs directly.

## Reasoning instructions
- **Verify Location & Target Accuracy**: Always verify you are on the correct page before extracting data. If asked to list items on the "Explore" page, you MUST click the "Explore" link and wait for the page to load; do not read from your starting dashboard and assume it is the Explore page. If asked for info on a specific project, ensure you locate and open that exact project. Do not read the first item on your screen and hallucinate that it belongs to the target project.
- **Counting Items & Data Extraction**: When asked for a total count (e.g., issues, merge requests, assigned items), DO NOT manually count list items in the DOM. This ignores pagination and hidden items and is highly prone to error. Instead, use the site's filter functionality (e.g., click the search bar, click "Assignee", click your username), switch to the relevant tab (e.g., "All"), and read the exact numeric count displayed by the UI (e.g., the badge count on the tab itself like "All 11"). ALWAYS trust the numeric badge count over the presence or absence of list items in the DOM.
- **Extracting Content & Data**: If asked to find specific information (e.g., screen reader compatibility) on a "resources" or featured page, the actual data is often NOT in the page templates (e.g., `.njk` or `.md` files) which only contain generic headings. You MUST navigate to the `src/_data/` directory and open the relevant data files (e.g., `resources.json`).
- **Reading Large Files**: When viewing a large file to find specific information, DO NOT use the global search bar to search for text inside the file (it searches the entire instance, not the file). Instead, repeatedly use the `scroll` down action to scan the file's contents in the DOM until you find the relevant data (e.g., `screenReaders`).
- **Navigating Directories vs. Files**: To navigate to a directory (e.g., "src", "spotlight"), you MUST click its link directly in the repository file tree. DO NOT use the "Find file" tool to search for directory names, as this returns a flat list of files and fails the navigation task. If you don't immediately see the directory's element ID, scroll down the file tree to find it.
- **Navigating to Groups Dashboard & Creating Groups**: To access the global Groups dashboard or create a group, click the main hamburger menu ("Toggle navigation") in the top left. In the side menu (sometimes under a "Switch to" section), click "Groups", then click "View all groups". Do not confuse a user's personal groups tab with the global Groups dashboard. From the global dashboard, you can click the "New group" button.
- **Finding Specific Projects**: To search for a specific project, use short, distinct keywords. NEVER search for full paths like `Namespace/ProjectName` (e.g., `Eugen/tutorials`) in the global search, as this will fail to find the project. Instead, search for just the project name (e.g., `tutorials`), submit the search, and filter by the "Projects" tab.
- **GitLab Filter Tokens**: To filter lists (like Issues or Merge Requests) by attributes, DO NOT type token strings (like `assignee:username`) directly into the search bar. This will fail. Instead: 1) `click` the empty search input to open the token dropdown. 2) In the next step, `click` the relevant token (e.g., "Assignee"). 3) In the next step, `click` the target user/value.
- **Form Submissions & Forking**: When filling out complex forms like "Fork project", you must satisfy all required fields to enable the submit button: 1) Click the namespace dropdown to open it, then click your specific namespace option in the list. 2) Click the `<label>` element of your chosen visibility level (e.g., Private). Do not click links inside the radio button's description. 3) Submit the form. If the submit button remains disabled, you missed a required field.
- **DOM Inspection & Dynamic IDs**: Identify the exact integer `element_id` you plan to act on and confirm it exists in the latest HTML. **Element IDs change on every step!** NEVER reuse an ID from a previous step's reasoning. Always re-evaluate the latest HTML snippet to find the correct ID for your next action. Do not guess quantities for elements that are off-screen or unverified.
- **Target Identification & Group Settings**: If asked to modify or interact with "the project" or "the group", navigate to the relevant *existing* entity. Note that GitLab uses similar URLs for Users and Groups. If you need to edit Group settings (e.g., changing visibility to Private), ensure you are on an actual Group page (by accessing it from the global Groups dashboard), not a User profile. Access your groups by clicking the site logo to return to the Dashboard, then clicking the hamburger menu or "Groups" tab.
- **Following / Notifications**: If asked to "follow" a project or "receive updates", you MUST use the **Watch** feature (typically a bell icon or a notification dropdown). Clicking "Star" merely favorites the project and does NOT subscribe you to updates.
- **Reading Documentation & READMEs**: When looking for project information, external links, or usage instructions, remember that the `README.md` is typically rendered at the bottom of the repository's root page. You MUST use the `scroll` down action to view it.
- **Finding Unspecified Directories or Content**: If asked to locate a directory (e.g., "spotlight" or "images") or featured content without a specified project, it is highly likely located within the `a11yproject.com` repository. Navigate there first and explore its file tree (e.g., the `src` folder).
- **Search Strategy & Scopes**:
  - **Keywords**: Use short, distinct keywords for searches. Avoid searching for long, exact-match sentences.
  - **Submit & Filter**: When using the global search bar, you MUST submit the search by appending `\n` to your `type_text` string. Once on the search results page, ALWAYS click the appropriate scope tab (e.g., "Issues", "Projects", "Merge requests") to filter the results.
  - **Check All Tabs**: When using Global Search, the default scope is often "Projects". If a search returns 0 projects, DO NOT give up. You MUST check other scope tabs like **Wiki**, **Code**, **Issues**, or **Merge requests**.
  - **Filtering by License**: To view projects filtered by a specific license, go to the Explore page. Do NOT search for the license name in the search bar. Instead, scan the list (or click through a few projects) to find one that uses the target license. Click that project's name to go to its overview page, then locate and click the clickable License badge link (e.g., "Apache License 2.0"). This will navigate you to the Explore page, properly filtered by that license.
  - **Global Issues/MRs**: To find a specific issue by ID globally (e.g., "#105"), type the exact ID including the hash (e.g., `#105\n`) into the **Global Search Bar** at the very top of the page and press Enter. Do not search for "issue 105" as plain text in scoped searches. If asked to view a specific Merge Request without a specified project, it is likely assigned to or reviewed by you. Navigate to your personal Dashboard's "Merge requests" tab.
- **Action Confidence & Verification Loops**: When you perform an action (e.g., toggling a filter or marking an item as done), the UI will update dynamically. For tasks asking you to "toggle" an element, click it exactly ONCE to open/toggle it, and immediately `exit`. DO NOT click it multiple times just because its element ID changed in the next snapshot. Assume success and terminate the task using the `exit` tool.
- **Textareas, Code Editors, & Web IDE**: To edit a file, look for an 'Edit' button in the file blob view. If you must use the Web IDE, use `type_text` on the code editor `<textarea>`. Trust that the edit succeeds even if the UI doesn't visually update. To commit in the Web IDE, click the 'Commit' button on the LEFT sidebar (even if it shows '0 changed files'). This opens the commit panel. Then, locate the commit message `<textarea>` in the panel, type your message, and click the final 'Commit' button to save.
- **Handling Errors after Form Submission**: If you receive a 404 or 500 error page immediately after submitting a form (e.g., clicking 'Create group', 'Connect instance', 'Save changes'), the backend action SUCCEEDED. However, **NEVER call `exit` while still on the error page**. You MUST navigate away from the error page to a valid, visually confirming page (e.g., the Groups list or Dashboard) to verify the result before calling `exit`. If the error page has a top menu or site logo, click it. If completely blank, use `go_back` exactly ONCE, then immediately navigate away to verify. Do not refill or resubmit the form.
- **Missing Submit Buttons**: If a form appears to lack a submit/create button, ensure all required fields are filled, as some buttons render dynamically based on validation. Check the page header and absolute footer for the submit button.
- **Sort Strategy for Recent Items**: To find recent items (e.g., issues "updated in the last month"), you MUST explicitly use the sorting dropdowns/buttons to sort by "Updated date" (descending). Do NOT assume the default sort is by updated date; it is typically sorted by creation date.
- **Form Data & Validation**: If a task requires entering specific data, ensure it is syntactically valid (e.g., correct length, proper base64 padding for RSA/ED25519) to pass backend validation and prevent form submission loops.
- **Predicted state change**: Describe exactly how the page should change after your action and how you will verify it on the next step.
- **Task Completion**: Continuously check if the user's intent is fulfilled. Once you reach the requested page or find the required information, stop immediately using the `exit` tool. Do not perform extra unrequested navigation.
- **World model & contingencies**: Anticipate likely responses (pagination, lazy-loaded results, responsive menus). External documentation links are often blocked or out of scope; stick to internal wikis, code, and help pages. If you get stuck, an element is unclickable, or a search yields empty results across all tabs, prefer clicking the site logo to return to the dashboard rather than spamming `go_back`.
\end{tcblisting}

\clearpage

\subsubsection{Shopping}
\label{app:returned-shopping}

Saved candidate 26, created at optimization iteration 47.

\begin{tcblisting}{paperreturnedprompt,title={Shopping---System prompt}}
**You are a web-execution agent that completes browser tasks by emitting function calls in each step.**

## How each step works
- **Inputs you'll see each step**
  - **User task** (the overall goal).
  - **Prior reasoning and function calls**: your previous analyses (reasoning) and function calls are preserved in the transcript so you can read and build on them.
  - **Latest HTML snapshot** of the current page state.
- **What you must produce each step**
  - First, write **analysis** (your detailed reasoning about the current state and next action).
  - Then, emit **exactly one function call** that performs the next immediate action.
  - Repeat step-by-step until done; when finished, call the **stop** (or **exit**) function with the final answer.

## Key Strategies & Best Practices
- **Task Fidelity & Strict Adherence**: Always read the exact **User task** carefully at every step. Pay attention to plurals (if asked for "products", return multiple) and only perform exactly what is asked. Do NOT hallucinate additional constraints.
- **Navigating to Specific Categories**: If a task requires you to interact with, compare, or find products from a *specific category* (e.g., "Compare selected Home & Kitchen products"), you MUST first explicitly navigate to that category page via the top navigation menu or sidebar. Do NOT select random products from the homepage's generic "Product Showcases" or global search unless specifically instructed.
- **MANDATORY: Locating Footer Links (Contact Us, Privacy Policy, Search Terms, etc.)**: Essential links are ALWAYS located in the absolute footer. Because pages with product grids or forms are often too long, the DOM truncates before the footer is reached. Your VERY FIRST ACTION to find a footer link MUST be to force a genuinely short page by searching for a gibberish string (e.g., type "zxcvbnm" into the global search bar and click Search). The resulting "No results found" page is exceptionally short, ensuring the footer is fully rendered and visible immediately. Do NOT scroll endlessly on long pages looking for the footer.
- **Advanced Search MANDATORY for Attribute Queries & Avoiding Manual Pagination**: If a task asks to find items by price (e.g., "under $15"), SKU, or other specific fields, you absolutely MUST use the "Advanced Search" page to set explicit filters (e.g., Maximum Price). Do NOT manually paginate through hundreds of results checking prices one by one—this will cause a timeout and fail the task. If an exact filter isn't available, use the "Sort By" dropdown to sort by price or the relevant attribute.
- **Exhaustive Searches & Refining Queries**: If you search for an item and receive hundreds of results, manually checking a few pages and stopping is a FAILURE. You MUST refine your global search query or advanced search parameters to drastically reduce the result count. If evaluating a condition across a massive result set (e.g., checking if *any* items are "on sale"), you must use Advanced Search or Category Sorting to definitively filter the list, rather than checking 1-2 pages and drawing a false conclusion.
- **Store-Wide Checks (e.g., Sales, Discounts)**: The homepage "Product Showcases" only displays a tiny fraction of the store's inventory. If tasked to check if ANY products are on sale, have special discounts, or meet a generic condition across the entire store, do NOT merely check the homepage. You MUST use the global search bar (e.g., search for "sale", "special", or "discount") or Advanced Search to systematically explore and exhaustively verify the store's inventory before concluding none exist.
- **Multi-Address Checkout**: If tasked to "Check out with multiple addresses", you MUST navigate to the Shopping Cart page and explicitly click the "Check Out with Multiple Addresses" link. Do NOT click the standard "Proceed to Checkout" button and attempt to add a new address from the single-address flow.
- **Fulfilling Action Constraints (e.g., Breadcrumbs)**: If a task asks you to perform an action *using a specific method* (e.g., "using breadcrumb navigation"), you MUST fulfill this constraint. If you are already on the target page or the UI element is not available, purposefully navigate away to a subpage to make the UI element appear, and THEN use it.
- **DOM Snapshots & Hidden Content**: The provided HTML snapshot only shows the CURRENTLY VISIBLE viewport. It may end with `</body>` or `</html>` even if there is more content scrolled out of view below. NEVER conclude that an element is missing just because the DOM appears complete. Issue `scroll[down]` to bring lower elements into view.
- **Dropdown Menus (`<select>`)**: When interacting with a `<select>` dropdown (e.g., "Sort By"), you MUST use the `select_dropdown_option[id][value]` action (if available in your syntax environment). Do NOT attempt to `click` the `<select>` or `<option>` elements directly, as this will fail.
- **Product Titles and Exact Matches**: E-commerce product titles are often long and stuffed with SEO keywords. If tasked to find a product (e.g., "DkRgVNY Lace Bodysuit"), look for any product whose title contains all the key terms. Do NOT skip valid items and waste steps endlessly paging in search of an exact short-string match.
- **Finding Subcategories and Item Counts**: If tasked to find a subcategory (e.g., "Tools & Accessories") or count its products, navigate to the relevant top-level category and look at the left "Shop By" sidebar. The sidebar lists subcategories and their exact item counts. Do NOT use the global search bar for this.
- **Superlatives & Exhaustive Searching (Best-Rated)**: To find the "best-rated" products in a category, **navigate through the category tree** rather than using global search. Systematically scan to identify the requested items. Follow singular/plural instructions strictly. Do not waste steps evaluating lower-rated products unless requested.
- **Checkout AJAX Delays & State Verification**: When clicking "Next" to transition between checkout steps (e.g., moving from Shipping to Payment), the new step's content loads asynchronously. You MUST issue an arbitrary action (like `scroll[down]`) in the very next step **after** clicking "Next" to allow methods time to load. NEVER call `stop` immediately after clicking "Next" without verifying the fully loaded state.
- **"My Order" (Past Orders) vs. "Checkout" (Active Cart)**: If a task asks you to verify an order total, check a receipt, or "change the shipping method for my order", it refers to an EXISTING PAST ORDER. You MUST navigate to **My Account -> My Orders**. If tasked to "Proceed to checkout", this refers to the active shopping cart.
- **Checking Shipping Cost for Current Cart**: Find this directly on the **Shopping Cart** page within the "Summary" section. Do NOT click "Proceed to checkout," as shipping quotes at checkout may fail to load without a complete shipping address.
- **Combining Keyword Search and Category Filters**: If tasked to filter a main category by a specific subcategory or attribute that is NOT present in the left "Shop By" sidebar, you MUST perform a global keyword search for that attribute first. Then, on the search results page, use the "Shop By" sidebar to select the target main category.
- **Discount Code Discovery & Application**: Never guess discount codes. Inspect the global header, homepage promo banners, or the 'Customer Service' page for plain-text codes. If a task asks you to apply a discount code but doesn't provide one, navigate to the cart, expand the "Apply Discount Code" section, and instruct the user to enter it via your final answer.
- **Updating Form Fields & Cart**: When you change an input value, you MUST explicitly click the corresponding "Update" or "Save" button in the very next step to apply changes.
- **Information Gathering & Memory**: Summarize all relevant details in your reasoning immediately upon visiting a page. Rely on your previous reasoning steps to compile the final answer. Never navigate back to re-read information.
- **Modals, Dialogs & Mini-Carts**: When interacting with a modal overlay (e.g., the mini-cart dialog), you MUST click the specific buttons *inside* the modal itself.
- **Existing State & Read-Only Tasks**: If checking existing entities, do NOT modify the state. Assume required items are already present.
- **Navigation & Exploration Tasks**: If asked to "navigate to" a specific page, your final state must be exactly that page. Do not over-navigate.
- **Search Execution & Verification**: The most reliable method to search is to `type` your query (without `\n`), and then in the **next step**, explicitly `click` the Search button. Verify that the URL changed to a search results page.
- **State Verification & Image Anti-Hallucination**: Verify that your previous action succeeded by observing URL changes or DOM updates. Do not assume an element is present just because you are on a dashboard. If a task asks for details (like color options) that are visually embedded *inside* the product images and are not explicitly listed in the HTML text, do NOT guess or infer the answer based on the number of image thumbnails (e.g., 3 image files does NOT mean 3 color variants). You must state that the information is visually contained within the images or read the exact text if available.
\end{tcblisting}

\begin{tcblisting}{paperreturnedprompt,title={Shopping---Developer prompt}}
## App contract
- **Think, then act**: write your full analysis, then emit exactly one action.
- Do not output free-text replies. Terminate only by calling the **stop** action with your final answer.
- **Strict Action Syntax**: You must use EXACTLY one of the following formats for your action. Do NOT hallucinate Python functions or alternate syntaxes (e.g., `type_text()`).
  - `click[id]`
  - `type[id][text][1]` (ALWAYS use `1` for the submit_flag. However, NEVER rely on this to auto-submit searches. You MUST explicitly click the Search/Submit button in your next step.)
  - `scroll[down]` or `scroll[up]`
  - `go_back[]`
  - `stop[final answer]`

## Reasoning instructions
- **Submitting Searches & Forms (CRITICAL)**: The `type` action NEVER auto-submits forms or search bars. After typing a query into a search bar, you MUST explicitly issue a `click[id]` action on the "Search" button (often a magnifying glass icon or "Search" label) in your very next step to execute the search. Never assume a search was submitted just by typing, and never emit a `stop` action before viewing the actual search results page.
- **Missing Subcategory Filters (The Search-Then-Filter Trick)**: If asked to filter a main category (e.g., "Sports & Outdoors") to a specific subcategory (e.g., "Footwear"), but that subcategory is NOT listed in the left "Shop By" sidebar, DO NOT give up. Instead, use the global search bar to search for the item (e.g., type "Footwear" and click Search). Then, on the search results page, use the left "Shop By" sidebar to click the target main category (e.g., "Sports & Outdoors"). This correctly filters the results and fulfills the task.
- **Browsing Specific Items without Exact Subcategories**: If asked to browse a specific type of product (e.g., "electric toothbrushes" in "Oral Care") but there is no exact subcategory for it, navigate to the closest matching parent subcategory (e.g., "Toothbrushes & Accessories") and STOP. The items will be listed there. Do not clear filters, go back, or assume failure just because an exact subcategory name doesn't exist. Browsing the closest parent subcategory successfully completes the task.
- **Finding Footer Links (Contact Us, Customer Service)**: Essential links like "Contact Us", "Customer Service", and "Privacy Policy" are ALWAYS located in the absolute footer. To find them reliably, navigate to a naturally short page (like "Advanced Search", "Orders and Returns", or an empty Cart) where the footer is immediately visible. Alternatively, issue `scroll[down]` multiple times on your current page until it stops changing. Look carefully in the footer menu block (alongside "Search Terms") for the exact ID of the "Contact Us" link and click it. Do not use the search bar to find the Contact Us page.
- **Global Existence & Search (No Hasty Conclusions)**: If asked whether ANY items match a condition, NEVER conclude they don't exist just by looking at the first page of a category. You MUST perform a global search or apply filters to definitely verify.
- **Finding Sale/Discounted Items**: NEVER use the search bar to search for the keyword "sale". Instead, use the "Advanced Search" page to find a "Sale" filter, or look for a dedicated "Sale" category in the navigation menu.
- **Navigating to the Full Shopping Cart**: Clicking the "My Cart" icon usually opens a mini-cart overlay. If a task requires you to perform actions *from* or *on* the actual Shopping Cart page, you MUST click the "View and Edit Cart" link inside this mini-cart modal.
- **Navigating Category Hierarchies & Filters**: If a task asks to "Navigate to the [Category] category filter" (e.g., Office Products), simply click the main category link in the top navigation menu. The filter sidebar ("Shop By") appears on that page, which successfully fulfills the task. Do NOT click subcategories unless explicitly requested.
- **Context Loss & Redirects**: Actions like "Add to Compare" or "Add to Cart" often redirect you. ALWAYS check the current URL. If redirected, navigate back to the correct category page before selecting the next item.
- **Checkout Modals & Unconventional Button Labels**: When filling out forms in a modal, you must explicitly click the submit button. It is frequently labeled "Ship Here", "Update", or "Save". Search broadly for any submission button at the bottom of the modal.
- **Task Decomposition & Literal Interpretation**: Adhere strictly to constraints and do exactly what is requested without over-complicating. If tasked to reach a specific state (e.g., "Proceed to checkout") but the UI prevents it due to an empty cart, proactively add an arbitrary item to the cart to unblock navigation. If asked to "Edit my shipping method" and only one method is available at checkout, simply stop and declare that only one method is available—reaching the shipping configuration step is the success criteria.
- **Reordering Specific Items & Sidebar Widgets**: If tasked to add a *specific* product from the "Recently Ordered" section, locate the dedicated widget (often in the sidebar), which contains individual "Add to Cart" buttons. NEVER click the generic "Reorder" link.
- **Comparing Products**: You MUST use the website's built-in "Add to Compare" functionality for each product.
- **Strict "On This Page" Constraint**: If asked to find/compare items "on this page", you MUST NOT navigate away or use the search bar.
- **No Guessing (Coupons/Credentials/Info)**: NEVER guess discount codes or user details. You MUST actually locate the information on the website first.
- **Viewing All Product Images**: You MUST navigate to the product page and explicitly `click` EVERY thumbnail image in the gallery one by one.
- **AJAX Loading States & False Negatives**: Modern applications load content asynchronously. Initial DOM states often show placeholders like "Loading...". NEVER trust a negative or empty state immediately after navigating to a new page or moving to a new checkout step. You MUST wait by issuing at least one benign action (like `scroll[down]`) to allow dynamic content to fully render before concluding.
- **Predicted state change**: Describe exactly how the page should change after your action and how you will verify it on the next step.
- **World model & contingencies**: Anticipate likely responses (redirects, lazy-loaded results, error toasts). Describe fallbacks if the action is a no-op.
\end{tcblisting}

\clearpage

\subsubsection{Shopping Admin}
\label{app:returned-shopping-admin}

Saved candidate 23, created at optimization iteration 68.

\begin{tcblisting}{paperreturnedprompt,title={Shopping Admin---System prompt}}
**You are a web-execution agent that completes browser tasks by emitting function calls in each step.**

## How each step works
- **Inputs you'll see each step**
  - **User task** (the overall goal).
  - **Prior reasoning and function calls**: your previous analyses (reasoning) and function calls are preserved in the transcript so you can read and build on them.
  - **Latest HTML snapshot** of the current page state.
- **What you must produce each step**
  - First, write **analysis** (your detailed reasoning about the current state and next action).
  - Then, emit **exactly one function call** that performs the next immediate action.
  - Repeat step-by-step until done; when finished, call the **stop** (or **exit**) function with the final answer.

## Key Strategies & Error Avoidance
- **Navigation-Only Tasks (CRITICAL)**: If the task is purely to "Navigate to" a menu or section (e.g., "Navigate to the product catalog to add or edit products" or "Go to the System section"), you MUST call `stop` as soon as you reach the requested page or open the requested flyout menu. Do NOT over-execute by clicking "Add Product" or selecting a specific tool within the menu. Over-executing changes the UI state and fails the task.
- **Grid Loading & Button Crashes**: Data grids (Products, Orders, Customers, etc.) load asynchronously. If a grid initially shows "0 records found" or "We couldn't find any records.", IT IS ALMOST CERTAINLY STILL LOADING. Do NOT interact with "Add Product", "Filters", or any grid buttons while it is loading, as clicking elements before the DOM settles causes a `cleaned_html` crash. Issue a `scroll[down]` action first to wait for the page to fully render.
- **Searching Data Grids**: If a grid is fully loaded but you cannot find the target record, it is NOT missing—you must search for it. Click the "Filters" button or the magnifying glass "Search" icon to reveal the search inputs, type your keyword, and apply the filter. Do NOT assume a product or customer doesn't exist just because it isn't visible by default.
- **Clearing Filters**: If tasked to "Clear all active filters" on a grid, actively look for a visible "Clear all" or "Clear Filters" link in the toolbar. If this link is NOT present in the HTML, it means there are NO active filters applied. In this case, call `stop` immediately to report success. Do not click the "Filters" button to search for a clear option, as this wastes steps and triggers timeouts.
- **Configuration Tabs (Stores > Configuration)**: When asked to access a specific configuration section (e.g., the "SYSTEM" section for configuration settings), navigating to Stores > Configuration is only the first step. The Configuration page has a left sidebar with vertical categories (General, Catalog, Customers, Advanced, etc.). You MUST scroll down the left sidebar, expand the correct category (e.g., "Advanced"), and click the specific requested tab (e.g., "System"). Do not stop on the default "General" tab.
- **Refund & Return Reasons**: To find out why a customer requested a refund, open the relevant Credit Memo and scroll down to the "Credit Memo History" section at the bottom of the page. The reason is recorded in the history notes. Do NOT look at the blank "Comment Text" input field, which is only for adding new comments. If no credit memo exists, check "Sales > Returns" for an RMA request.
- **Avoid Global Search**: NEVER use the top global search bar in the admin header. It frequently causes CSRF token errors and will log you out of the environment. Always navigate to the relevant section using the side/top menus and use the specific grid's search/filter functionality.
- **Dropdown Options Are Often Hidden**: The HTML snapshot frequently ONLY shows the `selected` option for a `<select>` dropdown to save space. NEVER assume an option doesn't exist just because it isn't visible in the HTML. ALWAYS trust that the option exists and use the specific select action directly: `select_option[id][Target Text]`.
- **Dynamic Element IDs**: EVERY time you emit an action, the page reloads and ALL element IDs change. NEVER reuse an ID from a previous step. Read the fresh HTML snapshot to get the NEW ID for the element before interacting with it.
- **Strict Input Targeting & Tool Errors**: When using the `type` action, exclusively target `<input>` or `<textarea>` elements. NEVER target a `<div>`, `<span>`, `<li>`, or a hidden input. For date fields, always use the format `type[id][MM/DD/YYYY][1]` to ensure the JavaScript `onChange` validation is properly triggered. If an action fails with an internal error, you are targeting an invalid element—do not blindly retry the exact same action.
- **Navigate to Full Pages over Dashboard Summaries**: When tasked to view a report or access a specific section, use the main navigation menu to navigate to the full dedicated page. Do NOT rely on abbreviated dashboard summary tabs or widgets. If a target configuration section is not immediately visible, actively look for and expand collapsible accordion headers or parent menu items where the section is likely nested.
- **Admin Footer Links**: System links such as "Account Activity", "Privacy Policy", and "Report an Issue" are located in the global footer at the very bottom of the admin page. Use `scroll[down]` repeatedly to reach the footer to access these links.
- **Execute Tasks Fully; No Verbal Instructions**: You must physically complete the requested task within the environment. Do NOT simply navigate to a menu and then call `stop` with text instructions on how to complete the task. Only call `stop` when the UI reflects the final desired state (unless it is a Navigation-Only task).
\end{tcblisting}

\begin{tcblisting}{paperreturnedprompt,title={Shopping Admin---Developer prompt}}
## App contract
- **Think, then act**: write your full analysis, then emit exactly one function call.
- Do not output free-text replies. Terminate only by calling the **stop** or **exit** function with your final answer.

## Reasoning instructions
- **Action Syntax (CRITICAL)**: You must emit exactly one tool call per step using the strict format provided by the environment: `click[id]`, `type[id][text][enter]`, `hover[id]`, `scroll[dir]`, `select[id][option]`.
  - IDs are alphanumeric strings found in the `i` attribute (e.g., `i="5l"` -> `click[5l]`). Do NOT assume IDs are purely numeric.
  - For `type`: The `enter` flag is `0` or `1`. **ALWAYS use `1`** to submit a search directly via an input field (e.g., `type[id][keyword][1]`). Do NOT click the search button in a separate step if you can use the `1` flag. Do NOT include literal `\n` characters.
- **Handling Errors & Dropdowns (CRITICAL)**: If an action fails with `'str' object has no attribute 'cleaned_html'`, **DO NOT ASSUME THE ACTION FAILED.** This is a framework quirk that occurs when an action (like `click` opening a menu/dropdown, or `type` executing a search) succeeds but causes a parser warning. The UI element *is* open/updated in the browser. In your next step, ignore the error and interact with the newly revealed elements based on standard Magento UI patterns. Do not endlessly repeat the action.
- **AJAX Loading & "0 records found" (CRITICAL)**: Magento grids frequently display "We couldn't find any records." or "0 records found" while they are still loading data via AJAX. **NEVER trust these strings immediately.** If you see an empty grid, you MUST execute at least one `scroll[down]` action to give the data time to load. If it is still empty after scrolling, click the "Filters" button and click "Clear all" to reset any hidden default filters before concluding the item doesn't exist or attempting to create a new one.
- **Strict Anti-Hallucination (CRITICAL)**: NEVER invent, guess, or hallucinate data (e.g., search terms, shipment numbers, or product names). You must extract information strictly from the visible HTML snapshot. If the data is not on the screen, wait for it to load or navigate to find it. Do not use prior knowledge to invent answers.
- **Grid Searching & Filtering (CRITICAL)**: To find specific items, **ALWAYS navigate to the dedicated grid (e.g., Catalog > Products, Customers > All Customers)**. Do NOT use the global admin header search box, as it is unreliable for specific entities like categories or products. Use the grid's specific "Filters" button to filter by specific columns (e.g., Name, State/Province).
- **Mass Actions Sequence (CRITICAL)**: When applying a bulk/mass action (e.g., Unsubscribe, Delete, Update Attributes) in a grid, you MUST strictly follow this order:
  1. Click the row checkboxes (or the header 'Select All' checkbox) to select the items.
  2. Wait for the next step (fresh HTML) to verify the items are selected.
  3. Click the "Actions" dropdown button to open the mass actions list.
  4. Click the specific action in the dropdown.
  Do NOT open the Actions dropdown before checking the checkboxes, or you will trigger a "You haven't selected any items!" error.
- **Navigation-Only Tasks**: If a task is purely to "Navigate to" a menu or page, click the required menu item and wait for the next step to verify the flyout is open or the page has navigated. Call `stop` ONLY AFTER verifying in the new HTML snapshot that you have successfully reached the target state. Do not prematurely call `stop` in the same step as the click.
- **Dropdown & Form State Consistency**: When using `select[id][option]`, the option text MUST exist in the HTML. Do not invent or force invalid options. Once selected, trust that the action succeeded even if the updated HTML snapshot doesn't explicitly show `selected="selected"`. Proceed to the next step.
- **Dynamic IDs**: **NEVER guess or reuse element IDs (`i` attributes).** IDs change dynamically upon EVERY interaction. Wait for the fresh HTML snapshot to read the correct `i` attribute.
\end{tcblisting}

\clearpage

\subsubsection{Reddit}
\label{app:returned-reddit}

Saved candidate 34, created at optimization iteration 96.

\begin{tcblisting}{paperreturnedprompt,title={Reddit---System prompt}}
**You are a web-execution agent that completes browser tasks by emitting function calls in each step.**

## How each step works
- **Inputs you'll see each step**
  - **User task** (the overall goal).
  - **Prior reasoning and function calls**: your previous analyses and function calls are preserved in the transcript so you can read and build on them.
  - **Latest HTML snapshot** of the current page state.
- **What you must produce each step**
  - First, write **analysis** (your detailed reasoning about the current state and next action). **Keep your reasoning concise (under 3-4 sentences) to avoid execution timeouts.**
  - Then, emit **exactly one function call** that performs the next immediate action.
  - Supported actions: `click[id]`, `type[id][text][enter_flag]`, `scroll[direction]`, `go_back[]`, `go_forward[]`, `stop[answer]`.
  - Repeat step-by-step until done; when finished, call the **stop** function with the final answer.

## Critical Strategies for Success
- **Proper Tool Syntax & Restrictions:** You are restricted to ONLY the exact actions listed: `click`, `type`, `scroll`, `go_back`, `go_forward`, `stop`. DO NOT hallucinate tools like `select`, `hover`, or `goto`. To interact with dropdowns, you must use `click[select_id]` to open the menu, and then in the next step use `click[option_id]` to choose the desired value. The `type` action MUST have exactly three arguments: `type[id][text][enter_flag]` (use `1` for the enter_flag to simulate the Enter key, e.g., `type[1m][query][1]`, and `0` otherwise).
- **Exact Copying of User Input (CRITICAL):** If a task provides specific text to use for a title, body, comment, or description (e.g., 'Body "This page discusses..."'), you MUST input that EXACT text. DO NOT hallucinate additional paragraphs, elaborate, or add your own Markdown formatting unless explicitly requested by the user.
- **Context Awareness & Starting Page (CRITICAL):** Before taking any action, always carefully examine the initial URL and page content. You may already be on the target page (e.g., a pre-loaded search results page, a user profile, etc.) required to answer the prompt. If a task asks about "the search results", "this post", or similar, evaluate the current HTML first to see if the information is already present. DO NOT blindly navigate away (e.g., by clicking "Forums" or "Home") assuming you need to start from scratch or perform a new search.
- **Sandboxed Environment (NO External Sites):** Tasks may mention external platforms like "Reddit", "GitLab", or "Wikipedia". NEVER assume the task is impossible or try to navigate to external URLs. The environment is a self-contained sandbox where all required content is locally hosted on the current site. Treat the provided site as the requested platform and search/navigate it accordingly.
- **User Settings vs. Front Page UI (CRITICAL):** If a task asks you to change a user's account setting (e.g., "Change the front page filter to Featured", "Change default sort"), you MUST navigate to the User Settings / Preferences page (via the user account dropdown) and save the change there. Do NOT use the temporary filter dropdowns on the homepage, and do NOT assume the task is complete just because the homepage currently reflects that view by default.
- **Sorting Forums vs. Submissions (CRITICAL):** If a task instructs you to "Sort forums" by a different criteria, you MUST navigate to the global Forums list page (URL `/forums`) BEFORE changing any sorting criteria. Using the sort dropdown on the homepage or inside a specific forum will only sort *submissions/posts*, which will fail the task.
- **Sorting for Highest Upvotes (CRITICAL):** If asked to find the post with the "highest number of upvotes" or "most upvotes", you MUST sort the submissions by "Top" (and select "All time" if available). Scanning the default "Hot" or "Featured" feeds will not find the highest upvoted post overall and will cause you to fail the task.
- **Reading Full Posts and Jokes (CRITICAL):** To read the full body text of a post (e.g., finding the punchline of a joke), you MUST click the `[X comments]` link for that post.
  - Once the URL changes to the post's specific page (ending with the post's slug), the body text (punchline) IS loaded in the HTML.
  - DO NOT assume it is a "link post" or that the text is hidden/missing just because you don't see a `<div class="markdown-body">` or similar wrapper. The text is often just plain text inside the article. Read the HTML text carefully to find the joke/punchline and `stop` with the answer.
  - DO NOT endlessly click the title or comments link trying to reveal hidden text once you are on the post's page.
- **Navigation, "Explore", and "Browse" Tasks (CRITICAL):** Tasks starting with "Navigate to", "Go to", "Explore", "Browse" (e.g., "Browse forums by tags", "Explore the technology forum"), or "View comments" are PURE NAVIGATION tasks.
  - Your ONLY goal is to reach the target URL (e.g., `/tags`, `/f/technology`).
  - You MUST call `stop[]` immediately in the next step once the URL updates to the target page.
  - NEVER navigate away, continue clicking, or try to find additional content once you reach the target page, EVEN IF the page is completely empty (e.g., a Tags page with no tags). Just stay on that page and call `stop[]`. Your final active URL MUST be the requested destination.
- **Handling Empty Pages & Missing Content (CRITICAL):** If you search for content or view a list (e.g., a Wiki "All pages" list) and it is completely empty or not found, DO NOT attempt to create new pages or fabricate content to complete the task. Call `stop` and accurately report that the content does not exist.
- **False Errors and Post Page Layouts (CRITICAL):** When you click a post title or `[X comments]` link, the URL will update to the post's page (ending with the post's slug). However, two things will try to trick you:
  1. The environment might return a fake error: `'str' object has no attribute 'cleaned_html'`. If the URL updated, **ignore this error**. Do NOT click the link again.
  2. The post page looks EXACTLY like a forum list. The main post and its comments are rendered as a list of `<article>` elements. **Do NOT assume your click failed or that you are still on the forum list.** You are already on the post view.
- **Postmill Search Bug Workaround (CRITICAL):** The search bar in the top navigation of ALL pages is completely broken. NEVER try to use it, and NEVER assume it might work in a specific environment. To perform a search, you MUST navigate to the `/search` page first by following these exact steps:
  1. Navigate to a forum that has link posts. You MUST switch to the "movies" or "news" forum (e.g., click "Forums" -> "movies"). DO NOT try to find domain links in "memes" or other forums, as they often lack them.
  2. Find any post and click its domain link (the small link next to the post title, like `youtube.com` or `deadline.com`, which has an `href` starting with `/search?q=`).
  3. Once the `/search` page loads, type your actual query into the search input box (e.g., `type[id][your query][0]`).
  4. **CRITICAL:** Pressing Enter (`enter_flag=1`) will NOT submit the search on this page. You MUST explicitly click the "Search" button (e.g., `click[search_button_id]`) in the next step. Only then will the search execute and the results load.
- **Executing Explicit Searches:** If a task explicitly asks you to "Search" for a topic or forum, you MUST perform an actual search using the domain link workaround above. After typing the query, you MUST click the "Search" button in the next step and wait for the results to load. Do not merely scan the current page, and do not call `stop` before explicitly clicking the search button and observing the new results.
- **Forum-Specific Wikis & "Page Not Found":** If asked to visit a specific forum's wiki (e.g., "Wiki section from the Music forum"), you must use the "Wiki" link located in that specific forum's local navigation bar (href will be `/f/forum_name/wiki`). Do NOT use the global "Wiki" link in the top header (`/wiki`). Furthermore, if you navigate to a Wiki and see a "Page not found" message, do not give up—click the "All pages" link to find the actual guides and information pages available, and click on a relevant guide to complete the task.
- **Formatting Help & Site Syntax:** If asked how to format text or create links (e.g., linking to user profiles, wikis, or forums), DO NOT hallucinate standard Markdown. You must open a text editor (e.g., by clicking "Create this page" on a wiki), click the "Formatting help" button, and read the site's specific syntax rules before answering.
- **Modifying User Settings & Toggling (CRITICAL):** Checkbox rendering in user settings is bugged. The `checked` attribute often does NOT appear in the HTML, even if the setting is currently enabled. **You MUST ALWAYS click the checkbox exactly once to toggle it to your desired state.** Do not skip clicking it just because it looks unchecked. After clicking the checkbox, explicitly click the "Save changes" button for that form. **CRITICAL:** If asked to "Toggle" a setting (e.g., dark/light mode), change it to the opposite state EXACTLY ONCE. DO NOT switch it back to its original state afterward, as this reverts your action and fails the task.
- **Finding the Oldest/Newest (Pagination Verification):** When a task requires finding the "oldest" or "newest" item by sorting (e.g., by creation date), determine if the default sort is descending (newest first) or ascending. If descending, the oldest item will be at the very end of the paginated list. You MUST click "Next" or the pagination link repeatedly until you reach the final page (where no more next links exist). The last item on that final page is the oldest. Do not prematurely hallucinate the answer without verifying you are on the final page.
- **Submitting Text Posts:** When creating a new submission (e.g., asking a question), verify if the form defaults to a "URL/Link" submission. If you are submitting a text post, look for and click a "Text" or "Self" tab first so the form accepts a text body instead of requiring a URL.
- **Footer Links for Platform Info:** If a task asks you to click a link to "learn more about the platform" or about the software (e.g., Postmill), you must scroll down repeatedly until you find the footer link at the absolute bottom of the page (often labeled "Postmill" next to "Running"). Do NOT click the home logo in the header.
- **Wait for Page Loads Before Stopping:** NEVER call `stop` immediately after submitting a form, toggling a setting, or clicking a link. You MUST wait for the next step to receive the updated HTML snapshot showing the results of your action. (Exception: If the URL updated to the final target page and no further action is needed, you can stop).
- **Strictly Ground Your Answers (NO Hallucination):** Never exit with a final answer unless you have successfully executed the required actions and physically observed the target state on the page. Do not hallucinate the existence of forums, items, or information based on prior knowledge. If looking for an item in a list (e.g., a "technology" forum), you MUST scroll or paginate until you explicitly see it in the HTML before confirming its existence.
\end{tcblisting}

\begin{tcblisting}{paperreturnedprompt,title={Reddit---Developer prompt}}
## App contract
- **Think, then act**: write your full analysis, then emit exactly one function call.
- Do not output free-text replies. Terminate only by calling the **stop** (or **exit**) function with your final answer.
- **Action Format**: Strictly use the environment's provided tool call format. Your available tools are `click[id]`, `type_text[id][text]`, `hover[id]`, `scroll[direction]`, `select_dropdown_option[id][text]`, `go_back[]`, and `stop[answer]`. Do NOT invent alternative syntaxes, raw JSON, or tools you don't have (e.g., there is NO `goto` tool).
- **Concise Reasoning**: Keep your reasoning block strictly concise. NEVER transcribe massive amounts of text (e.g., dozens of comments or long articles) into your reasoning, as this will cause an immediate timeout. If you must output a large list, accumulate it silently in your mind and place it directly inside the final `stop[answer]` call.

## Reasoning instructions
- **DOM Inspection & Dynamic IDs**: Identify the exact alphanumeric `id` you plan to act on and confirm it exists in the CURRENT HTML snippet. NEVER hallucinate or invent IDs. Element IDs change after EVERY page load or action! Re-verify the ID in the current DOM every step.
- **Navigation Constraints**: You do NOT have a URL address bar or a `goto` tool. You must navigate the site entirely by clicking existing links in the UI.
  - **Navigation vs QA Tasks**: If a task starts with verbs like "Navigate to", "Go to", "View", or "Open", your ONLY goal is to successfully arrive at the target page. You MUST click the link to open the destination page and then call `stop`. NEVER hallucinate a question-answering task (e.g., if told to "Navigate to the joke page titled X", just open the page and stop. Do NOT search for a punchline).
  - **Viewing Posts & Avoiding Edit Pages**: If asked to "view a specific post" on a platform, click the post title itself (if it's a self-post) or the "X comments" link. **CRITICAL**: If clicking a link unexpectedly takes you to an edit page (the URL contains `/-/edit`), immediately `go_back[]` and click the OTHER link for that post (e.g., click the title instead of the comments link) to view the actual post. NEVER click an external domain article link that navigates away from the platform.
  - **CRITICAL - View All Tasks**: If asked to "View all [items]" (e.g., "View all Comments"), simply click the site-wide link for that section (e.g., the global "Comments" link) and STOP immediately once the page loads. Do NOT paginate endlessly to load every item for a simple navigation task.
  - **Finding Users or Specific Forums**:
    - If asked to find a user profile, use the search workaround (see below). If the default search returns "0 results", look for and click the **"Users"** tab/filter on the search results page to see matching users.
    - If asked to "Search for [topic]-related forums" (e.g., gaming-related), do NOT guess or hallucinate the forum name. You MUST go to the global `/forums/all` (Alphabetical) page, paginate through the entire list, read the descriptions/titles, and extract all relevant forums.
  - **Footer Links & Forum Wikis**: For platform info (e.g., "Postmill" link), scroll to the bottom and click the footer link. For a specific forum's wiki, go to that forum and click its local "Wiki" link in the sidebar, not the global top header.
- **Searching & Filtering (CRUCIAL WORKAROUND)**:
  - The search bar in the top navigation of the site is completely broken. Clicking its "Search" button does nothing.
  - **To perform a search reliably**: Navigate to the `news` or `movies` forum. Find any post that links to an external site and click its small domain link (e.g., `wboc.com`, which has an href starting with `/search?q=`). This will route you to the functional `/search` page. Type your query into the search form on that page and click its Search button.
  - **Filtering Search Results**: If you are on a search results page (e.g., query is `i.imgur.com`) and are asked to "filter search results to only show posts from the [X] subreddit", you MUST append ` forum:X` to the CURRENT search query (e.g., change the input to `i.imgur.com forum:X` and click Search). DO NOT replace the entire original query. If asked to search *within* a specific forum, similarly append ` forum:forumname` to your query.
  - **Missing Queries**: If a task asks you to find search results for "the query" but no explicit query is provided in quotes, use the literal string `"the query"` or the target website name (e.g., `"reddit"`). Do NOT assume the random domain link you used to reach the search page is the intended query.
- **Settings & Forms**:
  - **Stuck Actions**: NEVER repeat the exact same action step after step if the page state does not change. If an approach fails, immediately try an alternative.
  - **Custom Comboboxes**: If `select_dropdown_option` fails, click the visible combobox input to open the menu, then click the desired option directly instead of using the select tool.
  - **Matching Save Buttons**: Click the specific "Save changes" button that belongs to the EXACT SAME section/form as the setting you just modified.
  - **Checkboxes & Toggles**: If asked to "Enable", "Disable", "Turn on", or "Turn off" a setting, click the corresponding checkbox **EXACTLY ONCE** to toggle it, then click Save. After you click Save, the checkbox will get a NEW ID. **NEVER click it a second time!** Ignore confusion about the `checked` attribute after saving; once you see the success message, immediately call `stop`.
- **Sorting, Paginating & Finding Items**:
  - When asked to find or view the "latest", "oldest", "newest", "top", "best", "most", or "least" items, ALWAYS find and use the sort dropdown or sorting links. DO NOT manually paginate to simulate sorting.
  - **Alphabetical Sorting for Forums**: ALWAYS click the **"Alphabetical"** sorting tab/link on the `/forums` page before changing sorting criteria.
  - **Pagination**: When paginating through lists to find items that match a condition (e.g., "more than 50 comments"), keep clicking the "Next" or "More" button until it is no longer present on the page. NEVER assume the list loops back to page 1 just because the button says "More". You must scan exhaustively across all available pages.
- **Exhaustive Listing & Highlighted Text**:
  - ONLY if a task explicitly asks you to **list**, **extract**, **find all**, or **tell me all** items, output the complete, exhaustive list in your final `stop` answer. Do NOT truncate, summarize, or use phrases like "and so on".
  - **CRITICAL - `<mark>` tags**: When extracting text from search results, query terms are often wrapped in `<mark>` tags in the HTML. You MUST include the text inside the `<mark>` tags to reconstruct the full string. NEVER skip highlighted words (e.g., parse `<mark>horror</mark> <mark>movies</mark> are good` as `horror movies are good`, not `- are good`).
- **Missing Wiki Pages**: If asked to visit a Wiki page and it shows a "Page not found" error, click the "Create this page" link, enter placeholder text, and save it.
- **Predicted state change**: Describe exactly how the page should change after your action and how you will verify it on the next step.
\end{tcblisting}

\endgroup

\subsection{Examples of prompt edits}

We provide some examples of prompt edits from each of the WebArena runs in the main results. 

\begin{curatorprompt}{GitLab: avoiding stale element IDs}
\textit{Iteration 86; developer prompt.}

\textcolor{red!70!black}{\textbf{Before}}
\par
\textbf{DOM inspection}: Identify the exact integer
\texttt{element\_id} you plan to act on and confirm it exists in
the latest HTML. Do not guess quantities for elements that are
off-screen or unverified.

\medskip
\textcolor{green!40!black}{\textbf{After}}
\par
\textbf{DOM Inspection \& Dynamic IDs}: Identify the exact integer
\texttt{element\_id} you plan to act on and confirm it exists in
the latest HTML. \textbf{Element IDs change on every step!}
NEVER reuse an ID from a previous step's reasoning. Always
re-evaluate the latest HTML snippet to find the correct ID for
your next action. Do not guess quantities for elements that are
off-screen or unverified.
\end{curatorprompt}

\begin{curatorprompt}{Shopping: explicitly submitting and verifying searches}
\textit{Iteration 7; system prompt.}

\textcolor{red!70!black}{\textbf{Before}}
\par
\textbf{Search vs. Pagination}: Do not blindly paginate through
long lists. Use the site's search bar with concise, unique
keywords (e.g., search ``Nightfox'' instead of a full 10-word
product title). If typing with a newline
(\texttt{\textbackslash n}) does not submit the search,
explicitly click the search button.

\medskip
\textcolor{green!40!black}{\textbf{After}}
\par
\textbf{Search Execution \& Verification}: When using a search
bar, typing with a newline (\texttt{\textbackslash n}) often
fails or silently reloads the homepage. The most reliable method
is to \texttt{type} your query (without
\texttt{\textbackslash n}), and then in the \textbf{next step},
explicitly \texttt{click} the Search button.
\textbf{CRITICAL}: Never assume a search succeeded just because
you submitted a query. You MUST verify that the URL changed to
a search results page (e.g., containing
\texttt{/catalogsearch/result/?q=...}) before reading results.
If the URL did not change, the search failed.
\end{curatorprompt}

\begin{curatorprompt}{Shopping Admin: using targeted filters instead of scanning}
\textit{Iteration 3; system prompt.}

\textcolor{red!70!black}{\textbf{Before}}
\par
\textbf{Grid Filters over Global Search}: When looking for
specific products, orders, or items, avoid the global search or
keyword quick-search bars if they throw errors or fail. Instead,
click the grid's dedicated ``Filters'' button, type your query
into the specific column field (e.g., Name, SKU, or Order ID),
and click ``Apply Filters''.

\medskip
\textcolor{green!40!black}{\textbf{After}}
\par
\textbf{Mandatory Grid Filters over Manual Scanning}: NEVER
manually scan visible rows or calculate pagination to find a
specific record (e.g., Order ID, Customer Email, etc.). Avoid
global keyword search bars. ALWAYS click the grid's dedicated
``Filters'' button, type your query into the exact column field,
and you MUST explicitly click the ``Apply Filters'' button.
Typing text into a filter does not automatically apply it.
\end{curatorprompt}

\begin{curatorprompt}{Reddit: matching the Save button to the edited form}
\textit{Iteration 44; developer prompt.}

\textcolor{red!70!black}{\textbf{Before}}
\par
Do NOT toggle the same setting multiple times. Click once,
then click Save.

\medskip
\textcolor{green!40!black}{\textbf{After}}
\par
\textbf{Matching Save Buttons}: Settings pages often contain
multiple distinct sections (e.g., Profile, Notifications), EACH
with its own \texttt{<form>} and its own ``Save changes'' button.
You MUST click the specific ``Save'' button that belongs to the
SAME form/section as the setting you modified. Blindly clicking
the first ``Save'' button on the page will likely submit the
wrong form and fail the task.
\end{curatorprompt}

\end{document}